\pdfoutput=1

\documentclass[11pt]{article}
\usepackage{tcolorbox}
\usepackage{xspace}
\tcbuselibrary{skins, breakable, theorems}
\usepackage{colortbl}
\definecolor{darkgreen}{rgb}{0,0.5,0} 
\definecolor{purple}{rgb}{1,0,1} 
\definecolor{todocolor}{rgb}{0.9,0.1,0.1} 
\definecolor{fixcolor}{rgb}{0.1,0.7,0.3} 
\definecolor{wycolor}{rgb}{0.9,0.1,0.1} 
\definecolor{hycolor}{rgb}{0.7,0.7,0.3} 
\definecolor{zwcolor}{rgb}{1,0,1} 
\newcount\DraftStatus  
\newcommand{\nbc}[3]{\ifnum\DraftStatus=1
	{\colorbox{#3}{\bfseries\sffamily\scriptsize\textcolor{white}{#1}}}
	{\textcolor{#3}{\sf\small$\blacktriangleright$\emph{#2}$\blacktriangleleft$}}
\fi}

\newcommand{\draftnote}[2]{\ifnum\DraftStatus=1
	\marginpar{
		\tiny\raggedright
		\hbadness=10000
		\def\baselinestretch{0.8}
		\textcolor{#1}{\textsf{\hspace{0pt}#2}}}
\fi}

\newcommand{\SystemName}{\textsc{CodeIP}\xspace}
\newcommand{\system}{\textsc{CodeIP}\xspace}

\usepackage[final]{acl}

\usepackage{graphicx}

\usepackage{times}
\usepackage{latexsym}
\usepackage{amsmath}
\usepackage{amssymb}
\usepackage{mathtools}
\usepackage{amsthm}
\usepackage{enumitem}
\usepackage{multirow}
\usepackage{subfigure}
\usepackage{booktabs}
\usepackage{arydshln}
\usepackage{algorithm}
\usepackage{algpseudocode}

\usepackage[T1]{fontenc}
\usepackage[utf8]{inputenc}
\usepackage{microtype}
\usepackage{inconsolata}

\title{\SystemName: A Grammar-Guided Multi-Bit Watermark for Large Language Models of Code}

\author{
Batu Guan$^1$~~Yao Wan$^1$\thanks{Corresponding Author.}~~Zhangqian Bi$^1$~~Zheng Wang$^2$~~\textbf{Hongyu Zhang}$^3$\\\textbf{Pan Zhou}$^1$~~\textbf{Lichao Sun}$^4$ \\
  $^1$Huazhong University of Science and Technology ~~~~ $^2$University of Leeds\\
  $^3$Chongqing University~~~~$^4$Lehigh University\\
  \texttt{\{batuguan,wanyao,zqbi,panzhou\}@hust.edu.cn}, \texttt{z.wang5@leeds.ac.uk}\\
  \texttt{hyzhang@cqu.edu.cn}, \texttt{lis221@lehigh.edu}\\
}

\begin{document}
\maketitle
\begin{abstract}
Large Language Models (LLMs) have achieved remarkable progress in code generation. It now becomes crucial to identify whether the code is AI-generated and to determine the specific model used, particularly for purposes such as protecting Intellectual Property (IP) in industry and preventing cheating in programming exercises.
To this end, several attempts have been made to insert watermarks into machine-generated code. However, existing approaches are limited to inserting only a single bit of information.
In this paper, we introduce \system, a novel multi-bit watermarking technique that inserts additional information to preserve crucial provenance details, such as the vendor ID of an LLM, thereby safeguarding the IPs of LLMs in code generation.
Furthermore, to ensure the syntactical correctness of the generated code, we propose constraining the sampling process for predicting the next token by training a type predictor.
Experiments conducted on a real-world dataset across five programming languages demonstrate the effectiveness of \system in watermarking LLMs for code generation while maintaining the syntactical correctness of code.

\end{abstract}

\section{Introduction}
\label{sec:intro}
Large Language Models (LLMs), particularly those pre-trained on code, such as CodeGen~\cite{nijkamp2022codegen}, Code Llama~\cite{roziere2023code}, and StarCoder~\cite{li2023starcoder}, have demonstrated great potential in automating software development. Notably, tools leveraging these LLMs, such as GitHub Copilot~\cite{friedman2021introducing}, Amazon's CodeWhisperer~\cite{amazon2023ai}, and ChatGPT~\cite{chatgpt}, 
are transforming how developers approach programming by automatically generating code based on natural language instructions and the context provided by existing code.
 
Although LLMs have shown significant potential in code generation, they also present challenges regarding the protection of Intellectual Property (IP) related to model architectures, weights, and training data, given the substantial costs associated with training a successful LLM~\cite{gpt-cost}. Furthermore, there are also increasing concerns about the use of generative AI in programming courses~\cite{bozkurt2023speculative}.
A crucial method for safeguarding the IPs of LLMs and detecting programming misconduct is 
to determine if a specific piece of code is generated by a particular LLM.

Watermarking~\cite{kirchenbauer2023watermark} offers a potential solution to determine the origin of machine-generated content. This technique is shown to be effective in Computer Vision (CV) and Natural Language Processing (NLP) domains.
It works by inserting information into multimedia formats (e.g., images and videos) without perceptibly diminishing the utility of content. 
By incorporating fingerprints such as owner/user ID, it supports leakage tracing, ownership identification, meta-data binding, and fortifying against tampering~\cite{mohanty1999digital}.

Existing watermarking techniques for LMs can be categorized into two groups: hard and soft watermarks.
A hard watermark is typically inserted by utilizing a masked language model like BERT~\cite{devlin2019bert} and RoBERTa~\cite{liu2019roberta} to replace tokens in generated content with synonyms. However, a hard watermark exhibits consistent patterns for different model inputs, compromising the protection strength. 
In contrast, soft watermarks are inserted during content generation, typically via manipulating the sampling probability distribution over the vocabulary during the decoding process of LLMs~\cite{kirchenbauer2023watermark}. As soft watermarks can adapt to the generated content, they change across model outputs, improving the diversity and strength of watermarks.

Recently, several attempts have been made towards watermarking LLMs for code generation, 
predominantly centered on two distinct approaches:
generating a one-bit watermark to discern the machine-generated nature of the code~\cite{lee2023wrote} or embedding a hard watermark through a semantic-equivalent transformation of the generated code~\cite{li2023protecting,sun2023codemark}. 
We argue that a single-bit watermark carries little information and is inadequate to preserve enough provenance information like the vendor ID of an LLM. 
Moreover, the implementation of a hard watermark does not offer robust protection~\cite{wang2023towards}, as the easily detectable nature of the hard-coded watermarking patterns undermines its effectiveness.

\begin{figure}[!t]
    \centering
    \includegraphics[width=\linewidth]{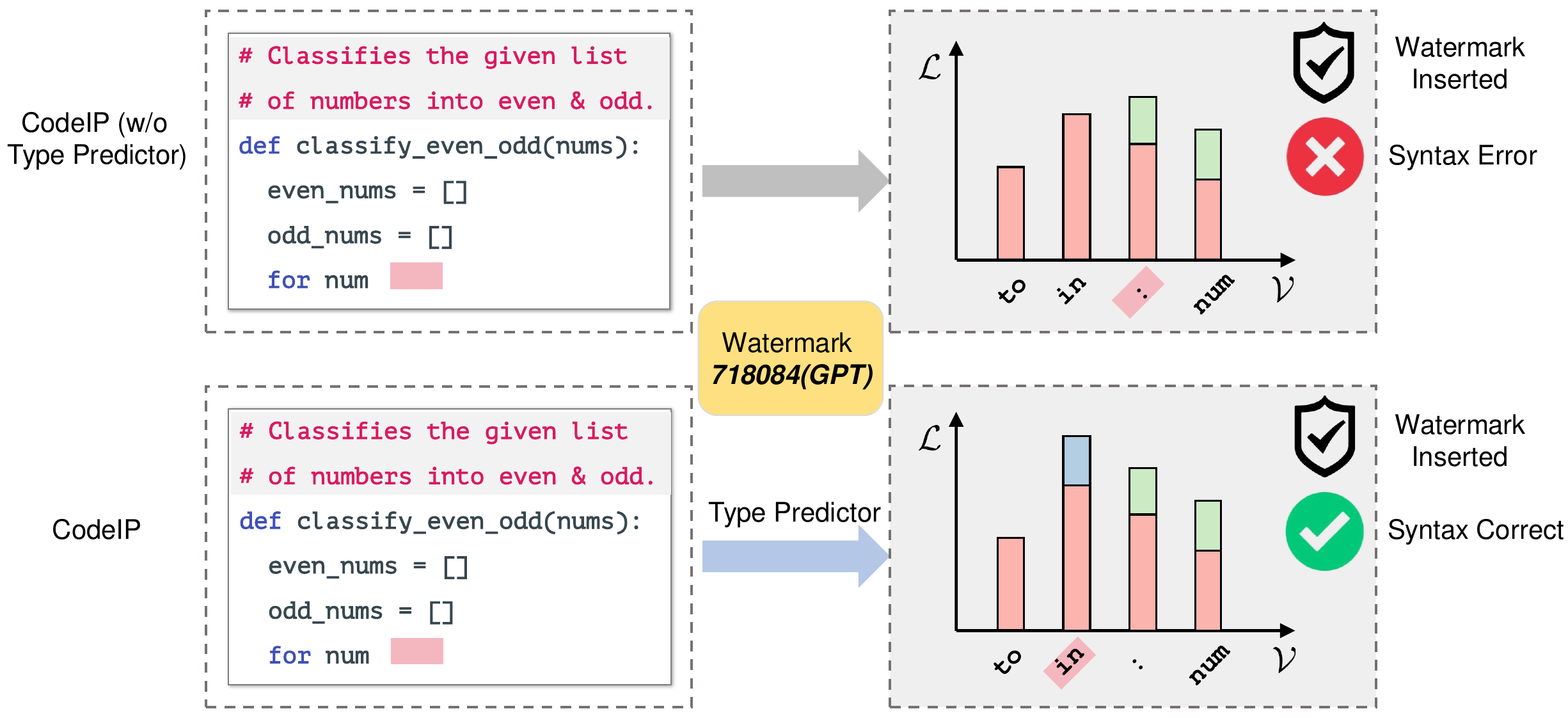}
    \caption{
    {\SystemName} can seamlessly embed multi-bit messages into LLMs while preserving the utility of the underlying code. ``\texttt{718084}'' is the ASCII value for ``\texttt{GPT}''.
    }
    \label{fig: motivation}
\end{figure}

To address the aforementioned limitations, this paper presents \system, a grammar-guided \emph{multi-bit soft watermarking} method for LLM-based code generation. 
\system inserts a watermark message based on the probability logits of LLMs during the code generation process, thereby embedding a multi-bit message in the generated code.
Moreover, \system incorporates grammar information into the process of generating watermarked code, maximizing the likelihood of generating semantically correct code.  This is achieved by training a type predictor to predict the subsequent grammar type of the next token, thereby enhancing the semantic correctness of the generated code.

Figure~\ref{fig: motivation} illustrates the advantages of type predictor introduced in \system.
In this example, our objective is to insert the multi-bit message (i.e., model name) ``\texttt{718084}'' (corresponding to the ASCII value of ``\texttt{GPT}'') into its generated code. 
Without grammar guidance, the LLM inaccurately predicts the next token as ``\texttt{:}''. 
However, the grammar analysis indicates that the succeeding token is expected to be a keyword. Our \system, which incorporates grammar constraints into the logit of LLMs, consistently tends to predict the correct token ``\texttt{in}''.
This capability preserves the semantic correctness of the code during the insertion of watermarks into LLMs.

We assess the performance of {\system} by inserting watermarks into code generated by three LLMs across five programming languages, namely Java, Python, Go, JavaScript, and PHP. 
Experimental results validate the efficacy of \system, demonstrating an average watermark extraction rate of 0.95. 
Importantly, our method preserves the utility of the generated code, achieving 50\% less CodeBLEU degradation compared to a baseline model without grammar constraints.

This paper makes the following contributions.
\begin{itemize}[left=0pt,itemsep=0pt]
    \item It is the first to study the problem of embedding the soft multi-bit watermarks into code LLMs during the code generation process.
    \item It presents a new method that utilizes the grammatical constraints of programming languages to guide the manipulation of probability logits in LLMs, thereby preserving the utility of watermarked code.
    
\end{itemize}
\paragraph{Data Availability.}
All experimental data and source code used in this paper are available at \texttt{\url{https://github.com/CGCL-codes/naturalcc/tree/main/examples/codeip}}~\cite{wan2022naturalcc}.

\begin{figure*}[!ht]
	\centering
	\includegraphics[width=0.98\linewidth,scale=1.0]
    {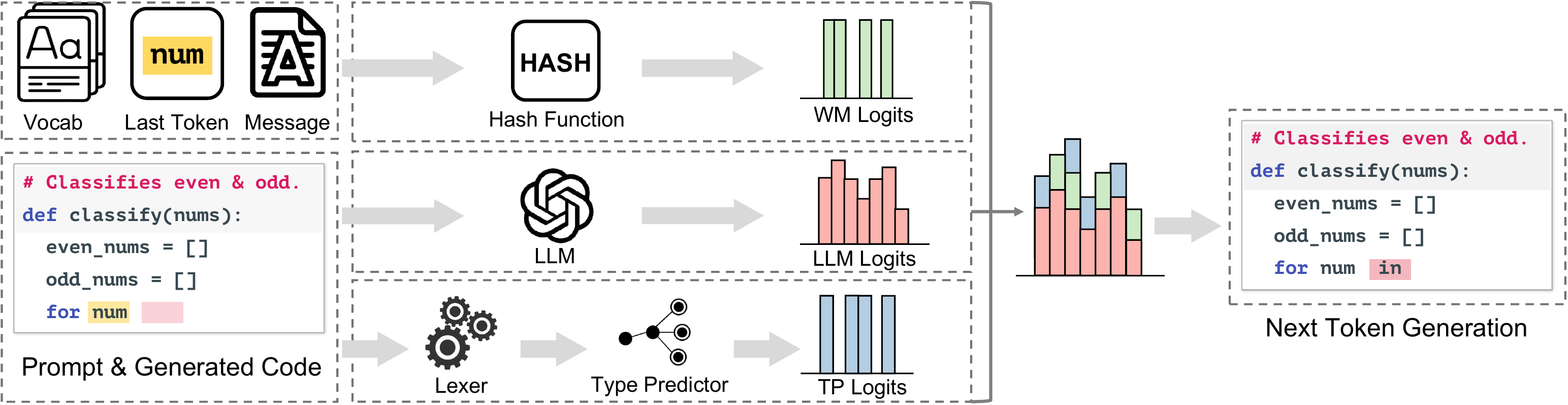}
	\caption{An overview of the watermark insertion process of \system. The last token of the generated code and the message are used to compute watermark logits (WM Logits) by a hash function. The prompt and the whole generated code are firstly used by LLM to compute LLM logits and consequently input into a lexer and a type predictor to compute type predictor logits (TP Logits). The three are added together to compute the final logits, which are used for decoding in the next token generation stage.}
\label{fig: Overview}
\end{figure*}
\section{Preliminary}

\subsection{Code Generation}
LLM-based code generation produces source code from high-level specifications or prompts.
Typically, these specifications (prompts) are conveyed through natural language descriptions, supplemented by partial code elements such as function annotations and declarations, which are provided by users.
Formally, let $\rho$ denote a prompt, which can be
tokenized into a sequence of tokens $\{w_{0}, w_{1}, \ldots ,w_{|\rho|}\}$, where $|\cdot|$ denotes the length of a sequence.
Let $\mathcal{V}$ denote the vocabulary used for mapping each token to corresponding indexes.
\begin{equation}
    p_{\text{LM}}(w_i) = \text{softmax}\left ( \mathcal{L}_{\text{LM}}(w_i|\rho, w_{0:i-1}) \right )\,.
\end{equation}
Here, $p_{\text{LM}}(w_i)$ denotes the probability distribution over the entire vocabulary $\mathcal{V}$, generated by the LM. 
We call the unnormalized score for each token in $\mathcal{V}$ produced by the LM as \textit{model logit}.
In this paper, the LM will always be an autoregressive Transformer~\cite{vaswani2017attention} pre-trained on source code, akin to the models in the GPT family, including Code Llama~\cite{roziere2023code} and StarCoder~\cite{li2023starcoder}.
Following this, the subsequent token $w_i$ is sampled from $p_{\text{LM}}(w_i)$ using specific sampling strategies, such as greedy sampling~\cite{10.5555/234285.234289} or multinomial sampling~\cite{bengio2000neural}.
In this paper, we adopt the greedy sampling strategy (cf. Appendix~\ref{app: sample}). Therefore, the next token will be sampled based on the following equation: $w_i = \arg \underset{w \in \mathcal{V}}{\max }  \, p_\text{LM}(w) $.

\subsection{The Problem: Watermarking the Code}
In this paper, our goal is to insert a multi-bit watermark message into a code snippet during the generation process of LLMs.
Typically, the watermarking algorithm comprises two stages: the insertion stage and the extraction stage.

During the process of inserting a watermark into the generated code, the initial consideration involves determining the specific message $m$ to be inserted as the watermark.
In practice, the model provider of an LLM can formulate a message, e.g. owner ID, to safeguard its model copyright.
It is noteworthy that while the initial content of message $m$ may encompass any characters, it undergoes conversion into a unique number before insertion.
Specifically, given the prompt $\rho$ and a watermark message $m$ as inputs, the $\text{INSERT}$ module produces a watermarked code $C = \text{INSERT}(\rho, m)$.

During the extraction stage, given an input snippet of code $C$, we expect that the module \text{EXTRACT} will produce its predicted watermark message $m^\prime = \text{EXTRACT}(C)$.

In the context of this formulation, the primary objectives of our watermarking for LLMs of code are twofold: 
1) to accurately insert the intent message as a watermark, and 2) to preserve the utility of the code without loss of semantics.

\section{\system}

This section provides a detailed description of \system. The {\system} comprises two distinct stages, namely insertion and extraction.
Initially, leveraging the decoding mechanism of existing LLMs, we use $\mathcal{L}_\text{LM}$ to denote the likelihood of each token in the vocabulary $\mathcal{V}$ to be inferred by the LLM. 
Subsequently, during the watermark insertion stage (cf. Sec.~\ref{subsec: pwm}), we incorporate the watermark message by calculating a logit value $\mathcal{L}_\text{WM}$ to influence the prediction choice of tokens. 
Moreover, we present a novel application of context-free grammar and introduce another logit at the insertion stage (denoted as $\mathcal{L}_\text{TP}$), which signifies the probability associated with the grammatical type of the subsequent token, to guide token generation from the perspective of grammar (cf. Sec.~\ref{subsec: ptp}). 
Finally, we integrate all the logits together (cf. Sec.~\ref{subsec: Combining the All}) and explain the watermark extraction techniques (cf. Sec.~\ref{subsec: detect}).

\subsection{Watermark Insertion}
\label{subsec: pwm}

The watermark insertion architecture is depicted in Figure~\ref{fig: Overview}. Following \citet{kirchenbauer2023watermark}, we insert the watermark into the generated code by modifying the probability distribution over the entire vocabulary $\mathcal{V}$ when LLM generates the next token.
We first select a set of tokens from the vocabulary using a hash function.
Based on the selected tokens, we compute the watermark logits, representing the likelihood of embedding the watermark message within each respective token.

\paragraph{Vocabulary Selection.}
The insight of inserting watermarks into code lies in selecting a set of tokens in the vocabulary under the control of the watermark message and enhancing their possibility of being generated during LLM decoding. 
We employ a hash function $\mathcal{H}$ to select tokens from the vocabulary $\mathcal{V}$.
Specifically, assuming that LLM is generating the $i$-th token and the previous generation is denoted as $[w_{0},w_{1}, \cdots, w_{i-1}]$, with watermark message represented by $m$. For any given token $w$ in $\mathcal{V}$, the hash function will take $(w,m,w_{i-1})$ as input and map it to either 0 or 1. 
We consider tokens $w$ that satisfy $\mathcal{H}(w,m,w_{i-1}) = 1$ as selected tokens, and our objective is to enhance their likelihood of being chosen by the LLM.

\paragraph{Watermark Logit.}
To augment the likelihood of generation, we calculate an additional logit referred to as the \textit{watermark logit} $\mathcal{L}_\text{WM}$ and incorporate it into the existing model logit $\mathcal{L}_\text{LM}$.
The implementation of the watermark logit $\mathcal{L}_\text{WM}$ relies on the outcomes of vocabulary partitioning. Assuming that the current LLM generates the $i$-th token $w_{i}$, preceded by the last token $w_{i-1}$, and denoting the watermark information as $m$, the watermark logit is computed as follows:
\begin{equation}
\label{eq: pwm} 
\begin{aligned}
\mathcal{L}_{\text{WM}}= \mathbb{I}\left(\mathcal{H}(w,m,w_{i-1}) = 1 \right)\,.
\end{aligned}
\end{equation}
Here, $\mathbb{I}$ denotes the indicator function.
By assigning a value of 1 to $\mathcal{L}_\text{WM}$ for those selected tokens whose resultant computation via the hash function equals 1, we can effectively enhance the likelihood of such tokens being preferentially chosen during the decoding stage of LLM.

\subsection{Grammar-Guided Watermarking }
\label{subsec: ptp}

Conventional watermarking methods, which randomly insert a message by perturbing the generation process for each token, often result in the disruption of the semantics within the generated code.
We posit that the generated code ought to adhere to the grammatical rules of the programming language. Consequently, we propose the integration of grammar constraints as a guiding principle in the code generation process. This inclusion is envisioned to maintain the utility of watermarked generated code.

\begin{figure}[!t]
	\centering
	\includegraphics[width=\linewidth,scale=1.0]
    {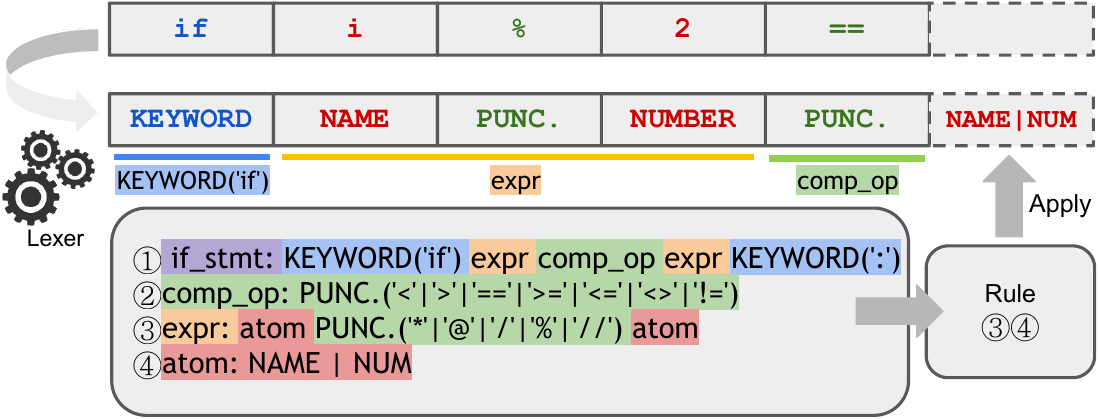}
	\caption{
 An example to highlight the role of CFG in ensuring the semantic correctness of generated code.
 }
	\label{fig: CFG}
\end{figure}

\paragraph{Context-Free Grammar (CFG).}
A CFG serves as a formal system for describing the syntax of programming languages, and possesses sufficient expressiveness to represent the syntax of most programming languages~\cite{hoe1986compilers}.
Typically, for a code snippet, a lexer, e.g. ANTLR~\cite{parr1995antlr}, can transform it into a sequence of lexical tokens. Subsequently, under the constraints of CFG rules, we can infer the potential type of the subsequent lexical token. For instance, as illustrated in Figure~\ref{fig: CFG}, after transforming the original code ``\texttt{if i \% 2 ==}'' into the sequence of lexical tokens, we can use CFG to infer the potential type of the subsequent lexical token as either ``\texttt{NAME}'' or ``\texttt{NUM}'', which could be helpful in the scenario of code generation.

Nonetheless, despite the constraints that CFG imposes on code, its direct application to the field of code generation still presents certain challenges. 
As demonstrated in the example of Figure~\ref{fig: CFG}, a CFG is capable of analyzing potential types for the subsequent lexical token. 
However, when multiple token types are considered valid next tokens, the utility of CFG in aiding code generation tasks becomes significantly limited, as it cannot calculate the probability distribution among these possible token types. Therefore, we train a lexical token-type predictor and intend to use it as a substitute for the CFG.

\paragraph{Lexical Token Type Predictor.}
We train a neural network to predict the lexical type of the next token. 
In particular, given the prompt and previously generated tokens, we initially employ a lexer to transform the given data into a sequence of lexical token types. Subsequently, this sequence is inputted into the predictor. The predictor then predicts a token type that will be outputted as the most probable lexical token type for the subsequent token.

In the context of LLM-based code generation, let \(\rho\) denote the prompt and \(G\) represent the generated code, assuming the LLM is currently generating the \(i\)-th token.
For any given code snippet denoted as $S = [\rho; G_{0:i-1}]$, where $[\cdot;\cdot]$ denotes the concatenation of two elements, it is feasible to extract its token sequence $T = \text{Lexer}(S) = [\tau_0, \tau_1, \ldots, \tau_{l-1}]$ via lexical analysis, where $\tau \in \mathcal{T}$ denotes the lexical token type and $l$ denotes the length of lexical token sequence.

Subsequently, an LSTM~\cite{hochreiter1997long} is adopted to serve as the type predictor and to predict the token type of the subsequent token by inputting the token sequence $T$, as follows:
\begin{equation}
\label{eq: lstm}
\begin{aligned}
    & \tau_{l} = \text{TP}(T)= \text{LSTM}(\tau_0, \tau_1, \ldots, \tau_{l-1})\,.
\end{aligned}
\end{equation}
Other neural networks, such as the Transformer~\cite{vaswani2017attention} can also be applied and we leave the exploration of other neural networks as our future work.

\paragraph{Type Predictor Logit.}
In order to mitigate the negative impact of watermarking on code utility, it is imperative to leverage our type predictor during the watermark insertion process, which is also the LLM decoding period. This necessitates transforming the predictive outcomes of the type predictor into a form of logit that can be added onto model logits. We name the new logit as \textit{type predictor logit}, which can also be represented as $\mathcal{L}_\text{TP}$.

The type predictor logits are scores of tokens within $\mathcal{V}$. Consequently, it becomes imperative to construct a dictionary in advance that associates each type of lexical token with potential LLM tokens corresponding to that particular type. For instance, the \texttt{KEYWORD} lexical token type encompasses LLM tokens such as ``\texttt{def}'', ``\texttt{if}'', and ``\texttt{else}'', while the \texttt{Punctuation} lexical token type incorporates LLM tokens including ``\texttt{(}'', ``\texttt{)}'', ``\texttt{;}'', ``\texttt{*}'', and so forth. We denote this dictionary by $\Phi: \mathcal{T} \mapsto \mathcal{V}$. Thus, $\mathcal{L}_{\text{TP}}$ can be calculated as:
\begin{equation}
\label{eq: ptp}
\mathcal{L}_{\text{TP}} = \mathbb{I}(w_i \in \Phi(\tau_{l+1})) \,.
\end{equation}
Finally, we can get into the process of generating the $i$-th token $w_i$.

\subsection{Putting it All Together}
\label{subsec: Combining the All}

Finally, the \(i\)-th token generated by the LLM can be formulated as follows:
\begin{equation}
\label{eq: total}
w_i = \arg\max _{w \in \mathcal{V}} \{\text{softmax}(\mathcal{L}_\text{LM} + \beta \mathcal{L}_\text{WM} + \gamma \mathcal{L}_\text{TP}) \} \,.
\end{equation}
Herein, $\beta$ and $\gamma$ represent hyperparameters for watermark logit $\mathcal{L}_\text{WM}$ and type predictor logit $\mathcal{L}_\text{TP}$.

\subsection{Watermark Extraction}
\label{subsec: detect}

In the watermark insertion stage, we employ $\mathcal{L}_\text{WM}$ to insert a watermark $w$ into the output $G$. Our strategy for watermark extraction involves enumerating all possible instances of the message (denoted as $m^\prime$), recreating the process of watermark insertion, and identifying the instance of $w$ that maximizes $\mathcal{L}_\text{WM}$, as follows: 
\begin{equation}
\label{eq: detect}
m_\text{ext} = \arg \underset{m^\prime}{\max }\left\{\sum_{i=1}^{L} \mathcal{L}_{\text{WM}}\left(w_{i} \mid m^\prime, w_{i-1}\right)\right\} \,,
\end{equation}
where $m_\text{ext}$ denotes the message extracted using Eq.~\ref{eq: detect} and $L$ denotes the length of token sequence in $G$. The insight of the extraction stage is that we determine that the most likely message causing the appearance of this generated code is the watermark inserted within this text. 

\begin{table*}[!t]
  \centering
  \small

  \setlength{\tabcolsep}{10pt} 
  \begin{tabular}{ccccccc}
    \hline
    \toprule
        LLM & Strategy & Java & Python & Go & JavaScript & PHP \\ 
    \midrule
    \multirow{2}*{Code Llama} & w/ WM + w/o TP & 0.90 & 0.93 & 0.87 & 0.98 & 0.97 \\ 
    ~ & w/ WM + w/ TP & 0.92 & 0.93 & 0.86 & 1.00 & 0.97 \\ 
    \midrule
    \multirow{2}*{StarCoder} & w/ WM + w/o TP & 0.88 & 0.98 & 0.90 & 0.97 & 0.96 \\ 
    & w/ WM + w/ TP & 0.86 & 0.97 & 0.87 & 0.96 & 0.96 \\ 
    \midrule
    \multirow{2}{*}{Deepseek Coder} & w/ WM + w/o TP &0.99 & 0.95 & 0.87 & 1.00 & 1.00 \\ 
    & w/ WM + w/ TP & 0.99 & 1.00 & 0.91 & 1.00 & 1.00  \\ 
    \bottomrule
\end{tabular}
\caption{The results of watermark extraction rate for different models with different strategies, where ``WM'' denotes Watermark and ``TP'' denotes Type Predictor. 
  }
  \label{table: rq1}
\end{table*}
\section{Experimental Setup}

\subsection{LLMs and Dataset}
To validate the effectiveness of our \SystemName, we choose three prominent LLMs: Code Llama \cite{roziere2023code}, StarCoder \cite{li2023starcoder}, and DeepSeek Coder \cite{bi2024deepseek} as our target models. 
We insert the watermark into the code generated by these selected models.
Note that, these models exist in different versions, each characterized by varying model sizes. In our experiments, we choose to employ the 7B model size, limited by the computation resources.
We select Java, Python, Go, JavaScript, and PHP from CodeSearchNet~\cite{husain2019codesearchnet} dataset and use the docstrings and function declarations as prompts. For each prompt, the LLMs generate the next 200 tokens.
Note that here we do not adopt HumanEval~\cite{chen2021evaluating} and MBPP~\cite{austin2021program} datasets as our evaluation datasets. This is because their code length is generally too short (cf. Appendix~\ref{app: dataset}) and not suitable for inserting watermarks. The relationship between the length of generated code and the extraction rate is studied in Sec.~\ref{subsec: rq3}.

\subsection{Implementation Details}
For all three LLMs, we implement a temperature of 0.75, a repetition penalty of 1.2, and no repeat n-gram size of 10. Given the distinct training processes of various LLMs, we establish the parameters $(\beta,\gamma)$ as $(5,3)$ for Code Llama and StarCoder, $(6,4)$ for DeepSeek Coder. We set the watermark message to be 2024 in our experiment and the whole possible watermark message set is $[0,2^{20}]$, which means we can insert a 20-bit message at most. 
The type predictor is an LSTM model, which encompasses an embedding layer characterized by an embedding dimensionality of 64. The hidden state dimensionality of the LSTM is 128. In our experiment, we train a type predictor for each language involved, given the distinct grammatical structures inherent to each language. 
We also compare the ability of \system to preserve code quality with other methods, particularly the vanilla methods from \citet{wang2023towards} and \citet{yoo2023advancing}. These methods have similar time consumption for inserting a watermark into a code snippet, making them suitable for comparison with our approach. The parameters for these comparative experiments have been adjusted to align with the code generation scenario.
All experiments are conducted on a Linux server with 128GB memory, with a single 32GB Tesla V100 GPU.

\subsection{Evaluation Metrics}
To evaluate the effectiveness of watermarking, one objective is to assess whether the watermark can be extracted from the generated code.
Specifically, we select 100 functions from the dataset for each programming language and extract the docstrings and declarations of each function to serve as prompts for LLMs generation.
We employ the extraction rate of watermarks as a metric to measure the efficacy of watermarking, reflecting the percentage of watermarks successfully extracted from the embedded code. Assuming there are $N$ prompts in the dataset, these $N$ prompts, when input to the LLM with \system, will generate $N$ segments of code with watermarks. By using \system, watermarks in $M$ segments of code are successfully extracted. The extraction rate will be calculated as follows.
\begin{equation}
\label{eq: detrate}
\text{Extraction Rate} = \dfrac{M}{N}\,.
\end{equation}
To validate the utility of watermarked code, we adopt the CodeBLEU~\cite{ren2020codebleu} metric, which has been widely adopted in the evaluation of code generation. The CodeBLEU metric can be depicted as follows. 
\begin{equation}
\label{eq: codebleu}
\begin{aligned}
    \text{CodeBLEU} & = \eta  \cdot \text{BLEU} + \lambda \cdot \text{BLEU}_\text{weight} \\ & + \mu \cdot \text{Match}_\text{ast} + \xi \cdot \text{Match}_\text{df}\,.
\end{aligned}
\end{equation}
Here, $\text{BLEU}$ is computed utilizing the conventional BLEU method as delineated by \cite{papineni2002bleu}. The term $\text{BLEU}_\text{weight}$ refers to a weighted n-gram match that is derived from juxtaposing hypothesis code and reference code tokens with varying weights. Furthermore, $\text{Match}_\text{ast}$ signifies a syntactic AST match which delves into the syntactic information inherent in the code. Lastly, $\text{Match}_\text{df}$ denotes a semantic dataflow match that takes into account the semantic congruity between the hypothesis and its corresponding reference.

In our experiments, we adopt the parameters recommended by \citet{ren2020codebleu} in their original paper, namely $(\eta, \lambda, \mu, \xi) = (0.10,0.10,0.40,0.40)$.
Note that, here we do not adopt the Pass$@k$ metric~\cite{chen2021evaluating}, which has been widely adopted to evaluate the LLMs for code generation. This is because the test cases are missing in our used CodeSearchNet dataset.
A detailed explanation of metrics is in Appendix~\ref{app: metrics}.

\section{Results and Analysis}
\subsection{Extraction Rate of Watermarks}
Table~\ref{table: rq1} shows a comparison among different kinds of watermarking strategies. 
Generally, under both watermarking strategies, the extraction rates consistently surpass 0.90 in most programming languages, indicating the efficacy of our watermarking techniques in the context of LLMs for code generation.
Taking DeepSeek Coder as an example, our watermarking strategy, both with and without the type predictor (``w/ WM + w/o TP'' and ``w/ WM + w/ TP''), demonstrates an impressive extraction rate of 0.99 for Java and 1.00 for PHP.
Moreover, the fact that the presence or absence of a type predictor has no obvious effect on the outcome is consistent with our initial expectations, as the type predictor is designed to prioritize the preservation of the utility of the generated code.

\subsection{Watermark \textit{vs} Code Quality}

\begin{table*}[!ht]
  \centering
  
  \small
  \setlength{\tabcolsep}{4.3pt} 
  \begin{tabular}{cccccccc}
    \hline
    \toprule
        LLM & Strategy & Java & Python & Go & JavaScript & PHP  \\ 
        \midrule
        ~ & Unwatermarked  & 28.99 & 22.56 & 31.73 & 23.01 & 44.56 & \\
        \cdashline{2-8}
        ~ & \citet{wang2023towards} & 23.76 (-5.23) & 10.04 (-12.52) & 21.52 (-10.21) & 16.32 (-6.69) & 32.85 (-11.71) & \\
        Code Llama & \citet{yoo2023advancing} & 26.87 (-2.12) & 8.91 (-13.65) & \textbf{28.29 (-3.44)} & 12.79 (-10.22) & 20.74 (-23.82) & \\
        ~ & w/ WM + w/o TP & 23.35 (-5.64) & 12.04 (-10.52) & 22.44 (-9.29) & 16.47 (-6.54) & 40.47 (-4.09) & \\ 
        ~ & w/ WM + w/ TP & \textbf{27.14 (-1.85)} & \textbf{12.25 (-10.31)} & 26.49 (-5.24) & \textbf{20.83 (-2.18)} & \textbf{40.61 (-3.95)} & \\ 
        
        \midrule
        ~ & Unwatermarked & 39.16 & 17.74 & 27.61 & 24.06 & 42.60 & \\
        \cdashline{2-8}
        ~ & \citet{wang2023towards} & 27.29 (-11.87) & 17.20 (-0.54) & 14.84 (-12.77) & 16.81 (-7.25) & 36.44 (-6.16) & \\
        StarCoder & \citet{yoo2023advancing} & 22.34 (-16.82) & 10.93 (-6.81) & \textbf{22.54 (-5.07)} & 15.08 (-8.98) & 17.36 (-25.24) & \\
        ~ & w/ WM + w/o TP & 25.70 (-13.46) & 17.60 (-0.14) & 13.39 (-14.22) & 15.25 (-8.81) & 40.11 (-2.49) & \\ 
         ~ & w/ WM + w/ TP & \textbf{32.11 (-7.05)} & \textbf{18.16 (+0.42)} & 17.55 (-10.06) & \textbf{19.18 (-4.88)} & \textbf{40.14 (-2.46)} & \\ 
        
        \midrule
        ~ & Unwatermarked & 32.10 & 19.68 & 33.10 & 23.97 & 42.29 & \\ 
        \cdashline{2-8}
        ~ & \citet{wang2023towards} & 24.84 (-7.26) & 15.95 (-3.73) & 26.38 (-6.72) & 19.61 (-4.36) & 36.84 (-5.45) & \\
        DeepSeek Coder &\citet{yoo2023advancing} & 21.78 (-10.32) & 17.62 (-2.06) & 26.56 (-6.54) & 17.18 (-6.79) & 18.09 (-24.20) & \\
        ~ & w/ WM + w/o TP & 25.55 (-6.55) & \textbf{18.35 (-1.33)} & 26.93 (-6.17) & 17.88 (-6.09) & \textbf{43.40 (+1.11)} & \\ 
        ~ & w/ WM + w/ TP & \textbf{31.22 (-0.88)} & 13.57 (-6.11) & \textbf{29.32 (-3.78)} & \textbf{19.65 (-4.32)} & \textbf{43.40 (+1.11)} & \\ 
        
        \bottomrule
    \end{tabular}
    \caption{CodeBLEU scores for different models with different strategies. The value in () represents the disparity in quality (CodeBLEU) between watermarked and unwatermarked code.}
  \label{table: rq2}
    \vspace{-1em}
\end{table*}

We further explore the impact of watermarking strategies on the utility of generated code.
Table~\ref{table: rq2} illustrates the overall performance of different LLMs when paired with different logits, measured by CodeBLEU.
From this table, it is evident that the use of watermark logit leads to a decrease in CodeBLEU scores for code generation across various models and languages, and with the subsequent incorporation of the type predictor logit, a distinct resurgence in CodeBLEU scores is observed across most settings. 
It should also be noticed that the performance of the logit predictor is superior to other similar works~\cite{wang2023towards, yoo2023advancing}. This emphasizes the significant efficacy of the type predictor in preserving the quality of code. 

\subsection{Parameter Analysis}
\label{subsec: rq3}

\paragraph{The Impact of Parameter $\beta$.} 
We conduct experiments on the variation in extraction rates when adjusting parameter $\beta$ under three LLMs. We only show the result of Java as an example in this section and more results can be seen in Appendix~\ref{app: beta}.
In Figure~\ref{fig: beta}, it can be seen that as $\beta$ continues to increase, the extraction rate of watermarks is also constantly increasing. When $\beta$ exceeds 5, an extraction rate of approximately 0.9 can essentially be achieved, which is relatively ideal. It indicates that watermark logit has a positive effect on whether watermarks can be extracted.

\begin{figure}[!t]
	\centering
	\includegraphics[width=0.42\textwidth]
    {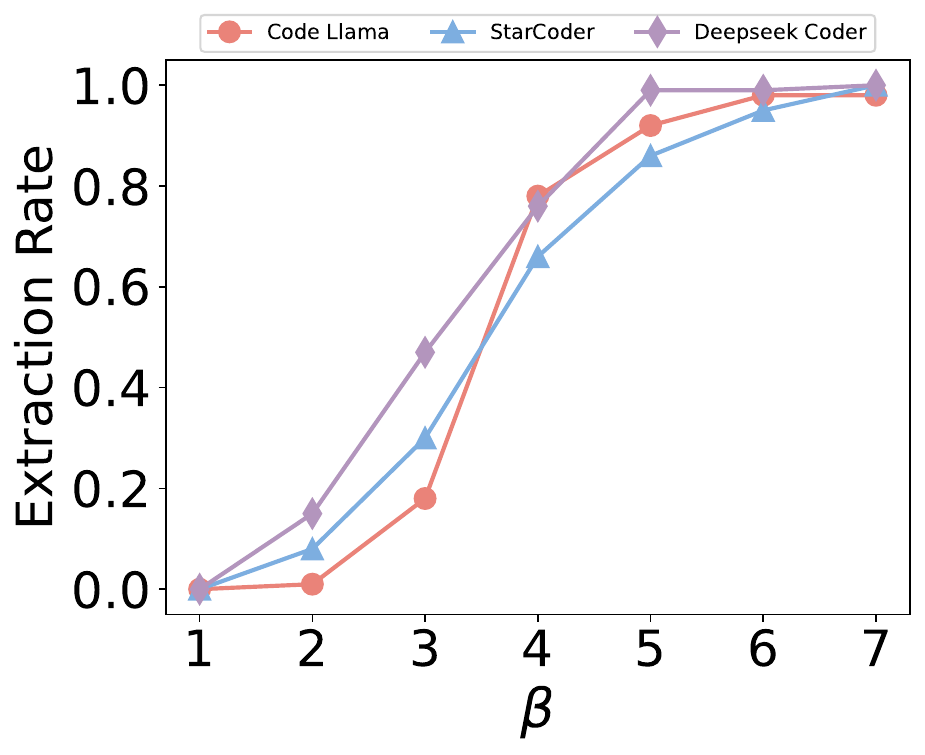}
	\caption{
    Impact of parameter $\beta$ on the extraction rate of generated Java code.
 }
	\label{fig: beta}
 \vspace{-1em}
\end{figure}

\begin{figure}[!t]
	\centering
	\includegraphics[width=0.42\textwidth]
    {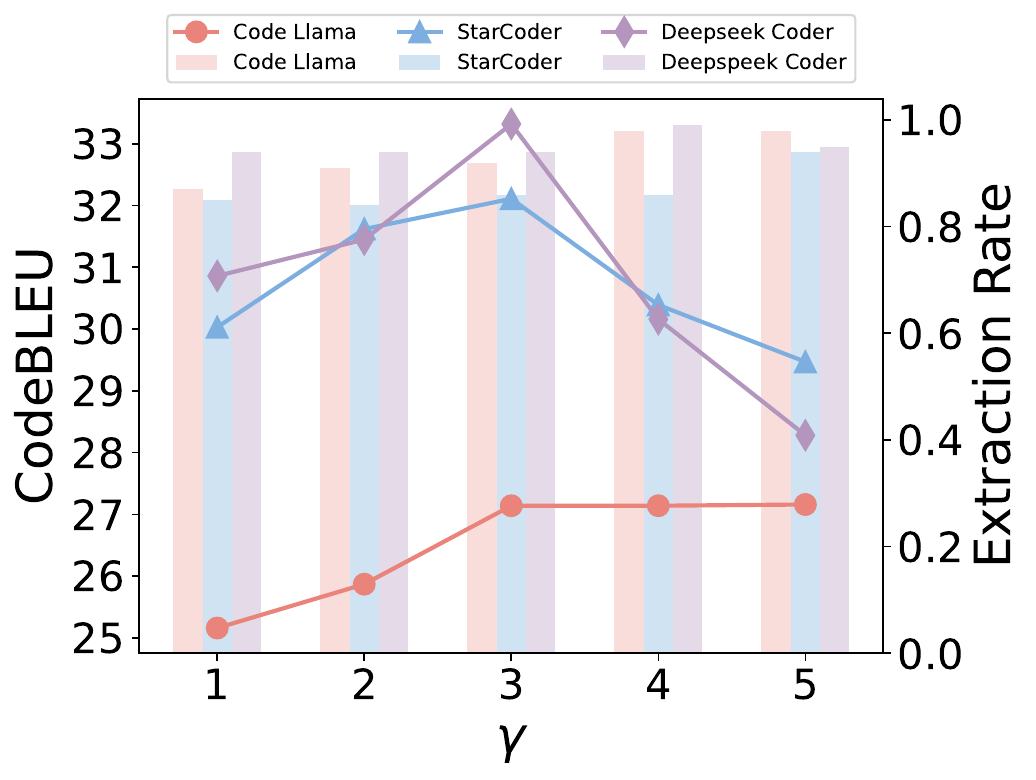}
	\caption{
    Impact of parameter $\gamma$ on CodeBLEU score and extraction rate of generated Java code. Lines are for CodeBLEU and bars are for extraction rate. 
 }
	\label{fig: gamma}
 \vspace{-1em}
\end{figure}

\paragraph{The Impact of Parameter $\gamma$.} 
We conduct experiments on three LLMs by varying parameter $\gamma$, aiming to investigate the impact of $\gamma$ on the CodeBLEU score and extraction rate of generated code.
The experimental results of Java, as depicted in Figure~\ref{fig: gamma}, reveal a noteworthy trend. The initial augmentation of $\gamma$ visibly improves the quality of the generated code. Nevertheless, as augmentation progresses beyond a certain threshold, a discernible decline in CodeBLEU becomes evident.
One plausible explanation for this inconsistency may stem from the inherent contradiction in tokenization, namely, the disparity between prevalent tokenization methods utilized by LLMs (e.g., WordPiece~\cite{schuster2012japanese} and BPE~\cite{sennrich2015neural}), and those employed by lexers.

For example, the LLM subtokens ``\texttt{ran}'' and ``\texttt{ge}'', when combined, can constitute the lexical token ``\texttt{range}'' which can be recognized during lexical analysis. Assuming the generated code to be ``\texttt{for i in ran}'', the subsequent LLM subtoken to be generated is most likely to be ``\texttt{ge}'', thereby rendering the generated code as ``\texttt{for i in range}''. 
However, from the perspective of a lexer, the token ``\texttt{ran}'' could potentially be classified as type ``\texttt{NAME}'', leading to the lexical token type being calculated as ``\texttt{PUNCTUATION}'', and thereby selecting ``\texttt{:}''.
Hence, the generation of code will be transformed into ``\texttt{for i in ran:}''. This contradiction caused by different segmentation methods between LLM tokenizer and lexical analysis can also lead to performance degradation when $\gamma$ is high.

Moreover, we measure the extraction rate under various $\gamma$ settings and observe that changes in $\gamma$ result in only minor fluctuations in the extraction rate. Thus, we conclude that the parameter $\gamma$ primarily affects the utility of the generated code, with minimal impact on the extraction rate.

\begin{figure}[!t]
	\centering
	\includegraphics[width=0.42\textwidth]
    {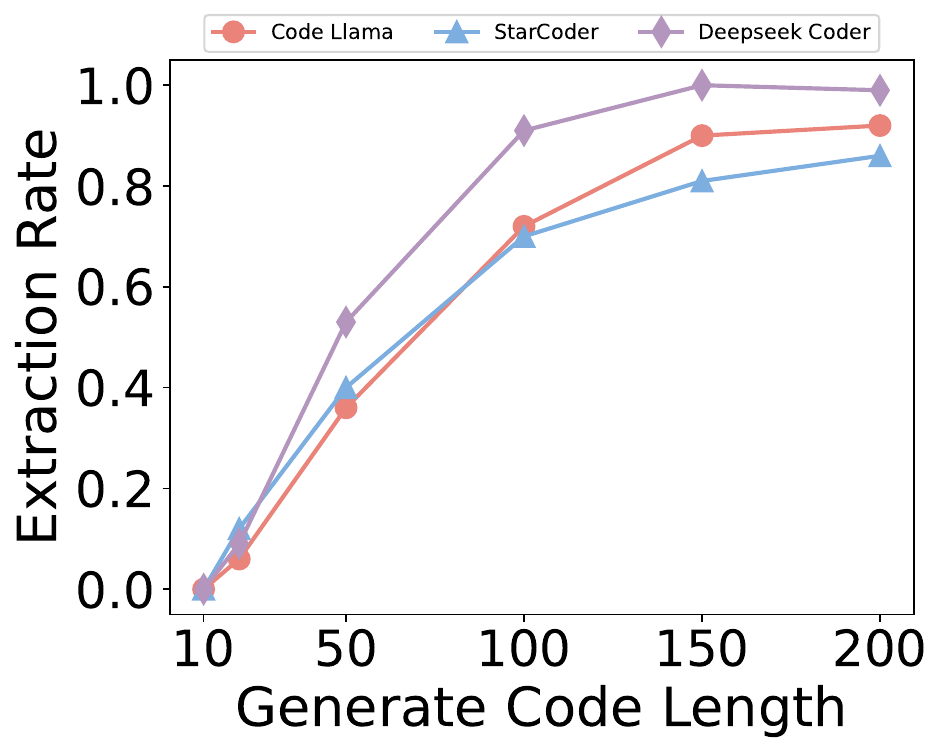}
	\caption{
    Impact of generated code length on the extraction rate of Java code.
 }
	\label{fig: gen_len}
\end{figure}

\paragraph{The Impact of Generated Code Length.} 
We also investigate the influence of generated code length, measured in terms of the number of tokens produced, on the effectiveness of watermark insertion. Our findings reveal a positive correlation between code length and the successful extraction rate, as depicted in Figure~\ref{fig: gen_len}. This observation underscores that the successful extraction rate of our watermark remains contingent on the length of the generated code. Specifically, shorter lengths of generated code lead to diminished distinctions between watermarked and unwatermarked code, consequently presenting a heightened challenge in extracting watermarks within such code.

\subsection{Resistance to Crop Attack} 
To underscore the robustness of our watermarking strategies, we consider a hypothetical scenario where developers use only a portion, rather than the entire generated code, to undermine the watermark—a situation termed a ``Crop Attack''. This involves subjecting the generated code to crop rates of 0.25 and 0.5, representing the removal of 25\% and 50\% of the code, respectively. The results are presented in Table~\ref{table: rq4}. Examination of the table reveals that, in most cases, our watermark's effectiveness only experiences a slight reduction under such rigorous attacks. These findings strongly indicate that our watermark exhibits notable resistance to crop attacks, demonstrating its 
robustness.

\begin{table}[!t]
\centering
\setlength{\tabcolsep}{4pt} 
\small
\begin{tabular}{cc|ccccc}
\hline
\toprule
LLM & Rate & Java & Python & Go & JS & PHP \\
\midrule
~ & 0 & 0.92 & 0.93 & 0.86 & 1.00 & 0.97 \\
Code Llama & 0.25 & 0.89 & 0.95 & 0.75 & 0.96 & 0.94 \\
~ & 0.50 & 0.71 & 0.85 & 0.51 & 0.87 & 0.87 \\
\midrule
~ & 0 & 0.86 & 0.97 & 0.87 & 0.96 & 0.96 \\
StarCoder & 0.25 & 0.81 & 0.95 & 0.85 & 0.93 & 0.95 \\
~ & 0.50 & 0.63 & 0.96 & 0.79 & 0.85 & 0.92 \\
\midrule
~ & 0 & 0.99 & 1.00 & 0.91 & 1.00 & 1.00 \\
DeepSeek Coder & 0.25 & 0.98&	0.99&	0.77&	0.94&	0.95\\
~&0.50 &0.91	&0.90&	0.56&	0.90&	0.87\\
\bottomrule
\end{tabular}
\caption{The performance of {\system} in code watermarking against crop attack.}
\label{table: rq4}
\vspace{-1em}
\end{table}

\section{Related Work}
\paragraph{LLM-based Code Generation.}
The roots of code generation can be traced back several decades~\citep{backus1957fortran, waldinger1969prow}.
Recently, many works focus on the intersection of deep learning and tasks of code~\cite{DBLP:journals/corr/abs-2401-00288}, such as code summarization~\cite{9000003,alon2018code2seq}, code search~\cite{10.1109/ASE.2019.00012}, code completion~\cite{li2024ircoco,sun2024sifting} and code generation~\cite{DBLP:conf/acl/Bi0W0GLZS0S24,DBLP:conf/coling/SunLL0ZLJ24}.
Currently, LLMs especially those pre-trained on code, such as DeepSeek Coder~\cite{bi2024deepseek}, Code Llama~\cite{roziere2023code}, CodeGen~\cite{nijkamp2022codegen}, StarCoder~\cite{li2023starcoder}, and CodeGeeX2~\cite{zheng2023codegeex}, have emerged as dominant forces in code generation. Leveraging the capabilities of these LLMs, several commercial tools are reshaping the programming landscape for developers, including GPT-3.5~\cite{chatgpt}, Gemini~\cite{gemini}, and GitHub Copilot~\cite{copilot}.

\paragraph{Software Watermarking.} 
The software watermarking problem has been studied since 1996 by \citet{davidson1996method}, who altered code blocks or operand order to insert watermarks. 
\citet{qu1998analysis} proposed a software watermark method based on graph coloring problem and graph structure of the code, which was further developed by \citet{myles2004software}, \citet{zhu2006recognition}. 
These rule-based early methods are often constrained by the usage scenarios and various attack techniques.

Recently, several works~\cite{yang2023towards, li2023protecting} have been focusing on watermarking the code generated by LLMs.
They utilized a post-processing approach, whereby watermarks are inserted through transformations applied to the code subsequent to its generation by the model. 
However, these techniques have several limitations, including their specificity to a single language and their vulnerability to counterfeiting once the watermarking method is disclosed, which restricts their applicability. Additionally, some multi-bit watermarking techniques~\cite{wang2023towards, yoo2023advancing} have been proposed, but these approaches tend to reduce the utility of the generated text.

\paragraph{Machine Generated Text Identification.} 

The task of identifying machine-generated text has always been of paramount importance. 
Early research focused on adding watermarks to arbitrary texts, while in recent years, studies on text watermarking have started their attempts to distinguish between machine-generated and human-generated texts.
An intuitive approach is to treat it as a binary classification task, accomplished by training a model~\cite{solaiman2019release,bakhtin2019real}. 
Another approach is to identify model-generated text by detecting features of the generated text. \citet{tay2020reverse} distinguished texts by detecting detectable artifacts in the generated text, such as sampling methods, top-$k$ probabilities, etc. There is also a dataset created for testing models ability to distinguish machine-generated text from human-written text~\cite{zhang-etal-2024-llm}.
In 2023, \citet{kirchenbauer2023watermark} introduced a novel method for embedding watermarks into text during model inference by altering the selection probabilities of certain tokens.
\citet{lee2023wrote} extended this method to code generation, incorporating threshold-controlled watermark inclusion.

\section{Conclusion}
In this paper, we propose {\system} to watermark the LLMs for code generation, with the goal of safeguarding the IPs of LLMs. 
We insert watermarks into code generated by the model, and introduce grammatical information into the watermark generation process by designing a type predictor module to safeguard the utility of generated code.
Comprehensive experimental findings affirm that {\system} exhibits a notable extraction rate, excels in safeguarding code semantics, and demonstrates a degree of resilience against attacks.
In our future work, we plan to persistently advance toward more secure LLM-powered software engineering through the continuation of our research.

\section{Limitations}
In our experiments, we adopt CodeBLEU for evaluation, which is a commonly used metric in assessing the quality of code generation. In our future work, we will employ additional evaluation metrics to assess the experimental results. 
We also strive to find datasets suitable for our work that can be evaluated using other metrics, such as Pass@$k$. 
Furthermore, the experiments have substantiated that our watermark exhibits a certain degree of robustness under crop attacks, as this is the most easily implemented attack method in model copyright scenarios. Other forms of attacks such as variable name obfuscation could potentially degrade the readability of generated code, thus making them less likely to be employed in attacks aimed at infringing model copyrights, which is an assault we aim to prevent. We will persistently investigate and enhance the robustness of our watermark to make it applicable for more protection scenarios.

\section*{Acknowledgements}
This work is supported by the National Natural Science Foundation of China under grant No. 62102157, and the Major Program (JD) of Hubei Province (Grant No. 2023BAA024). 
We thank all the reviewers for their insightful comments.

\nocite{langley00}

\bibliography{custom}

\appendix
\section{Lexical Token Type}
Despite the diversity in syntax among various programming languages, consistency remains at the lexical analysis level. That is, the types of tokens parsed out by lexical analysis are fundamentally similar. The text box below presents potential token types that may be parsed following lexical analysis.
\begin{center}
\fcolorbox{black}{gray!10}{\parbox{0.9\linewidth}{\texttt{`Token', `Comment', `Error', `Escape', `Generic', `Keyword', `Literal', `Name', `Operator', `Other', `Punctuation', `Text'}}}
\end{center}

\section{Learning the Type Predictor}
Formally, the type predictor accomplishes the task of the next lexical token prediction. We adhere to conventional training methodologies for this particular task to train it. For a given programming language, we postulate that the collected code dataset of this particular language is denoted as $\mathcal{D}$, and each segment of code within this dataset as $d \in \mathcal{D}$.
To facilitate the acquisition of pertinent language grammar by the type predictor, we initially employ a lexer specific to that language to transform each instance of $d$ into a corresponding lexer token sequence. Taking into account that the possible token type of the subsequent word is typically associated with the types of nearby tokens, for predicting the type of the $i$-th token, we extract $n$ preceding token types from the sequence to predict this $i$-th token type. Hence, for the dataset $\mathcal{D}$, our learning objective can be formulated as follows:
\begin{equation}
\label{eq: loss}
\mathcal{J}(\mathcal{D}) = \sum_{d \in \mathcal{D}} \sum_{i=n}^{|T_d|} \log p_\text{LSTM}(l_i | l_{(i-n):i})\,,  
\end{equation}
where $\mathcal{J}$ is the loss function utilized during the training of type predictor, and $|T_d|$ denotes the length of lexical token type sequence of original code $d$.

After training, each type predictor can achieve an accuracy rate of over 70\% on the test set.

\section{Detailed Explanation of Metrics}
\label{app: metrics}
\label{app: fpr}
In our experiments, we opt to assess the effectiveness of our watermarking system by detecting the extraction rate, rather than employing metrics such as FPR and AUROC. This is because, while one-bit watermark technology can only distinguish whether an article is generated by a machine, multi-bit watermarks can embed more information. Therefore, the ability to extract the information within is a more critical evaluation criterion.

In fact, based on the experimental results, we find that the false positive rate is 0, i.e., there is no record of successfully extracting watermarks from codes written by natural programmers.
Theoretically, it is also not difficult to see that extracting the correct watermark from non-machine-generated text is very challenging. Assuming there are $N$ possible watermark extraction results, for naturally generated text that is not machine-made, the results of watermark extraction can be considered to be uniformly distributed, hence the probability of extracting the correct watermark is $1/N$. In experiments, the total number of possible extraction results is set to $2^{20}$, under such circumstances, the probability of extracting the correct watermark is less than one in a million. Therefore, metrics such as FPR and AUROC are not suitable in our experiments.

\section{Dataset Analysis}
\label{app: dataset}
\begin{figure}[!t]
	\centering
	\includegraphics[width=\linewidth,scale=1.0]
    {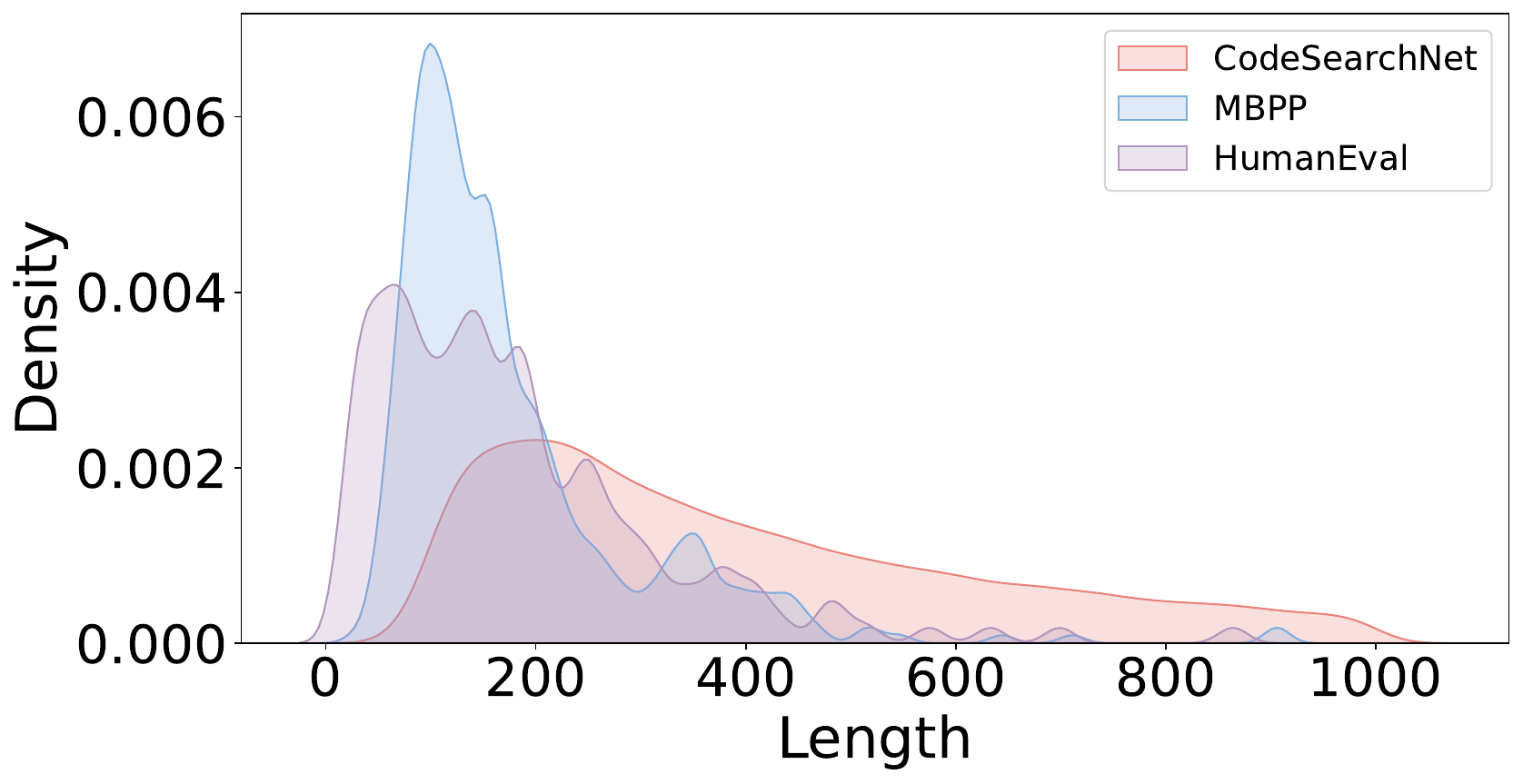}
    \caption{
    The length distribution of code in CodeSearchNet, MBPP, and HumanEval datasets. For better readability, code in CodeSearchNet exceeding 1000 characters has been truncated. The length is measured in characters.
    }
    \label{fig: dataset}
\end{figure}

In Figure~\ref{fig: dataset}, we examine the code lengths across three datasets: MBPP~\cite{austin2021program}, HumanEval~\cite{chen2021evaluating}, and CodeSearchNet~\cite{husain2019codesearchnet}. Analysis of both length distribution and average length reveals a notable distinction: the CodeSearchNet dataset exhibits significantly longer code lengths compared to the other two datasets.

\begin{figure}[!t]
	\centering
	\includegraphics[width=0.7\linewidth,scale=1.0]
    {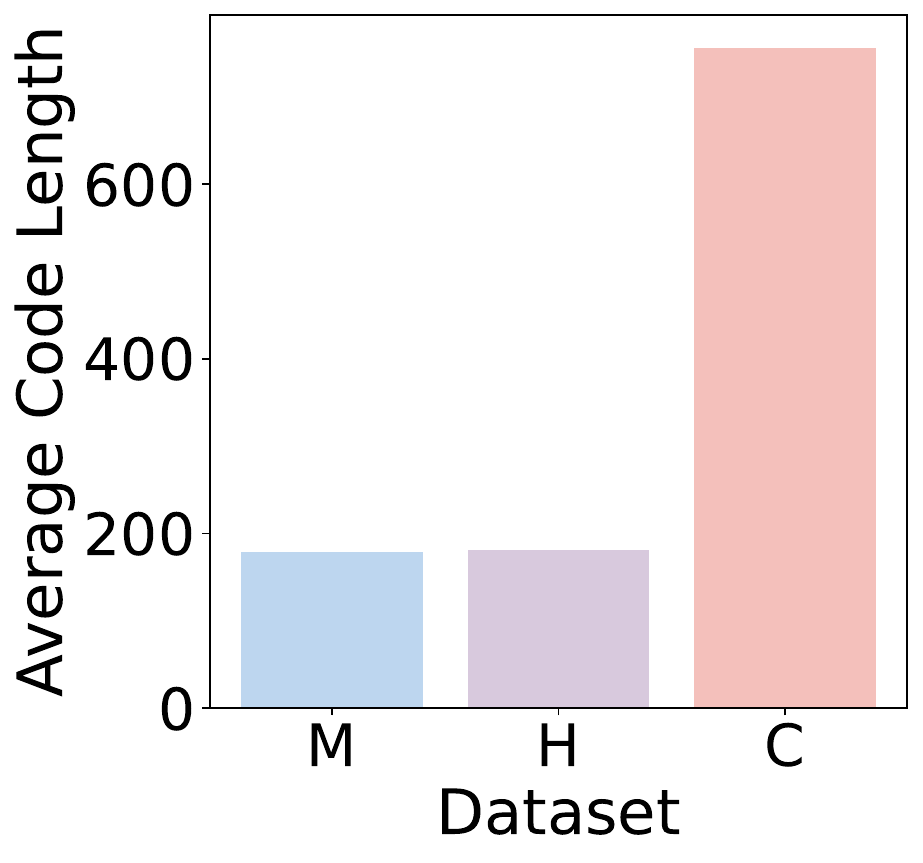}
    \caption{
    The average code length for CodeSearchNet, MBPP, and HumanEval, measured in characters. On the $x$ axis, ``M'' represents MBPP, ``H'' represents HumanEval, and ``C'' represents CodeSearchNet.
    }
    \label{fig: dataset_bar}
    
\end{figure}

\section{Sampling Strategy Selection}
\label{app: sample}
In our experiment we use greedy sampling as our decoding strategy. Other decoding strategies like beam search, top-p sampling, etc. can also be used. The selection of the decoding strategy will not affect our evaluation of \system. Due to the convenience of greedy sampling, we opt for this strategy.

\section{More Results on Parameters}

\subsection{Results of Parameter $\beta$}
\label{app: beta}
\begin{figure*}
	\centering
	\subfigure[Go]{
		\begin{minipage}[b]{0.23\textwidth}
			\includegraphics[width=1\textwidth]{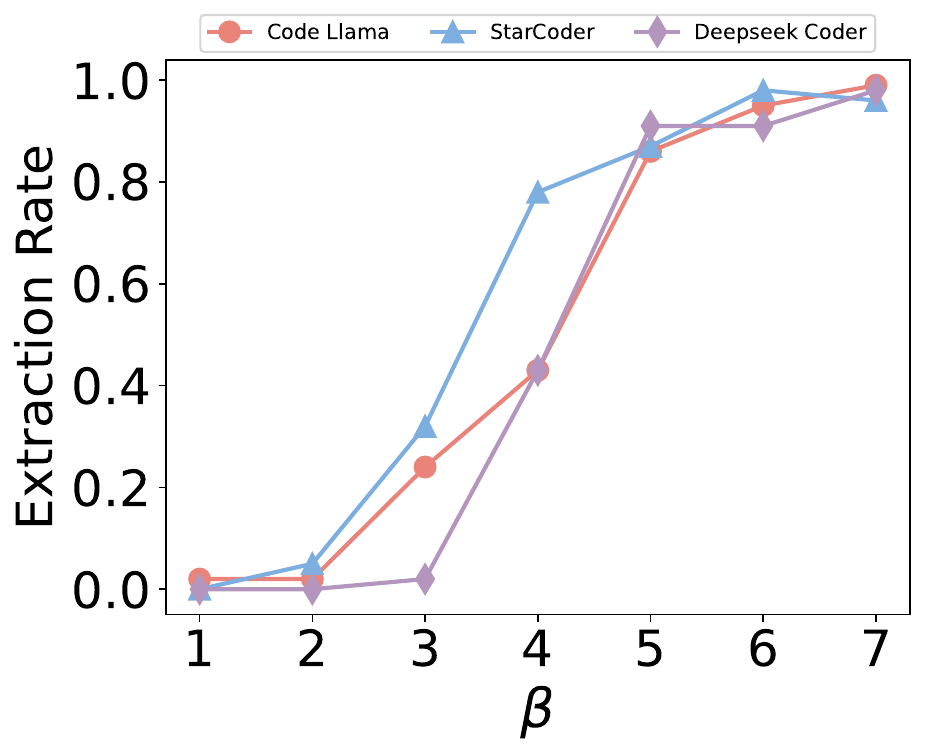} 
		\end{minipage}
		\label{fig:grid_4figs_1cap_4subcap_1}
	}
    	\subfigure[Python]{
    		\begin{minipage}[b]{0.23\textwidth}
   		 	\includegraphics[width=1\textwidth]{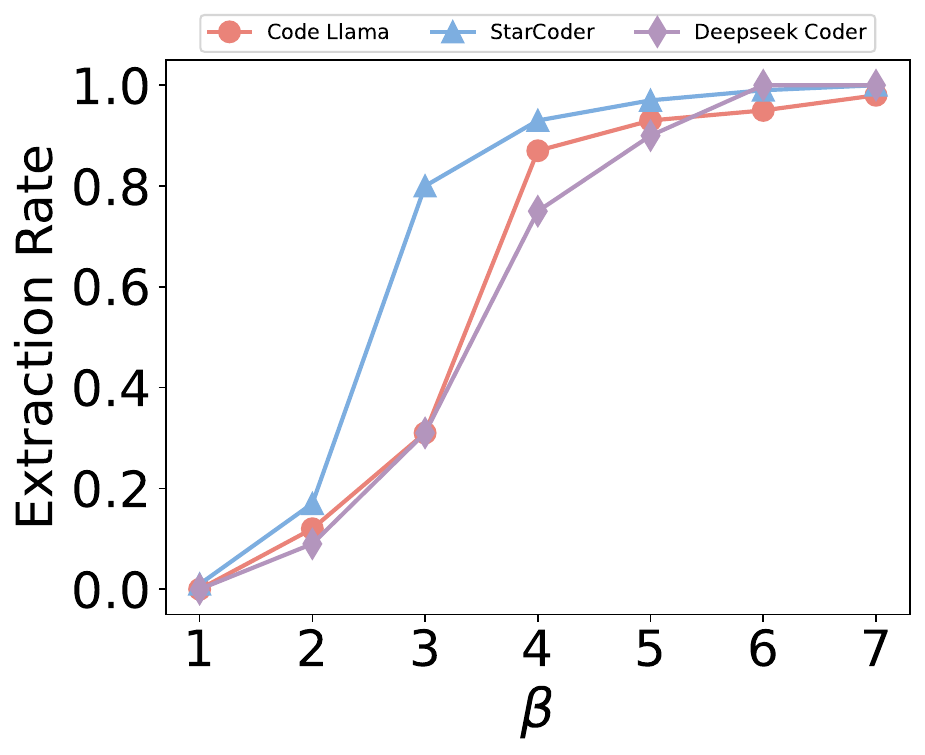}
    		\end{minipage}
		\label{fig:grid_4figs_1cap_4subcap_2}
    	}
	\subfigure[JavaScript]{
		\begin{minipage}[b]{0.23\textwidth}
			\includegraphics[width=1\textwidth]{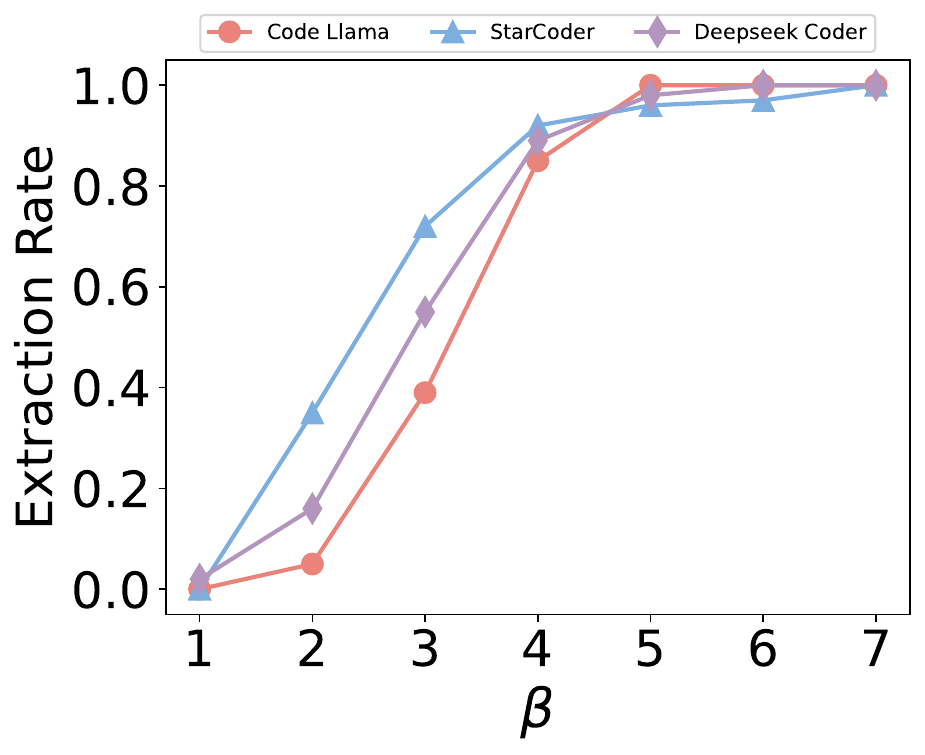} 
		\end{minipage}
		\label{fig:grid_4figs_1cap_4subcap_3}
	}
    	\subfigure[PHP]{
    		\begin{minipage}[b]{0.23\textwidth}
		 	\includegraphics[width=1\textwidth]{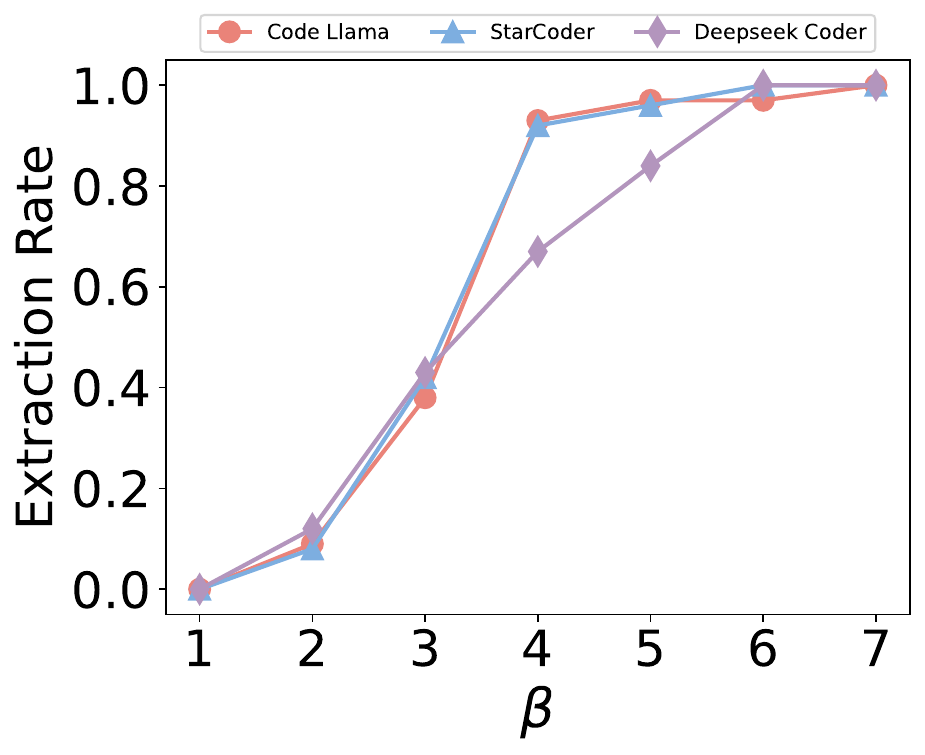}
    		\end{minipage}
		\label{fig:grid_4figs_1cap_4subcap_4}
    	}
	\caption{Impact of parameter $\beta$ on extraction rate of generated code.}
	\label{fig: app_beta}
\end{figure*}

We present additional results (Figure~\ref{fig: app_beta}) demonstrating variations in extraction rate as $\beta$ varies.

\subsection{Results of Parameter $\gamma$}
\label{app: gammaacc}
\begin{figure*}
	\centering
	\subfigure[Go]{
		\begin{minipage}[b]{0.23\textwidth}
			\includegraphics[width=1\textwidth]{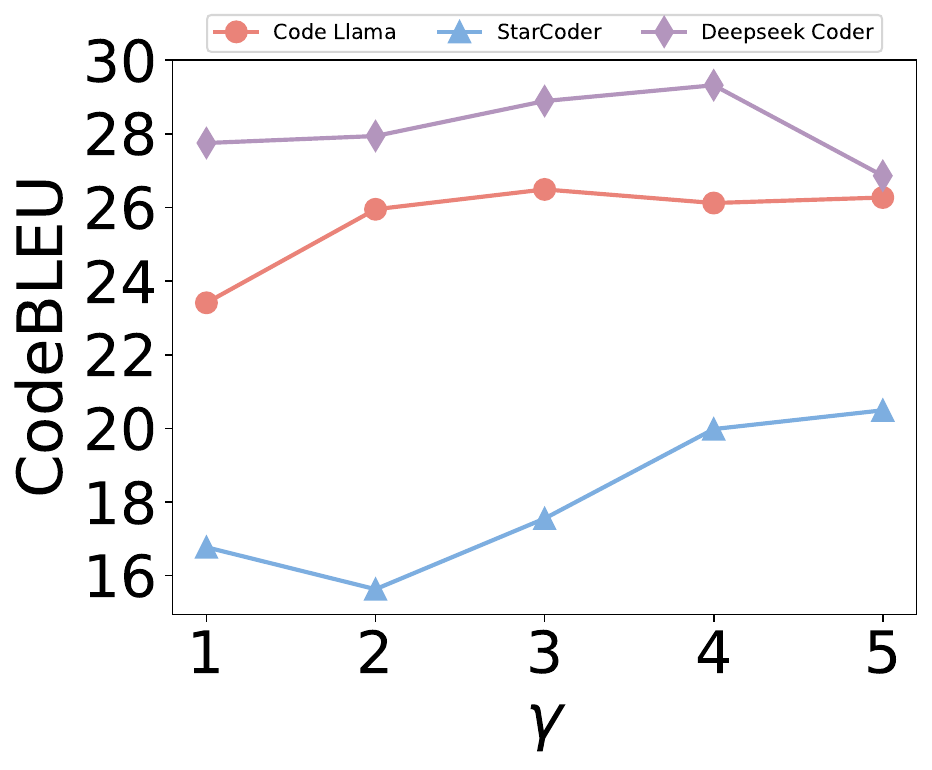} 
		\end{minipage}
		\label{fig:grid_4figs_1cap_4subcap_1}
	}
    	\subfigure[Python]{
    		\begin{minipage}[b]{0.23\textwidth}
   		 	\includegraphics[width=1\textwidth]{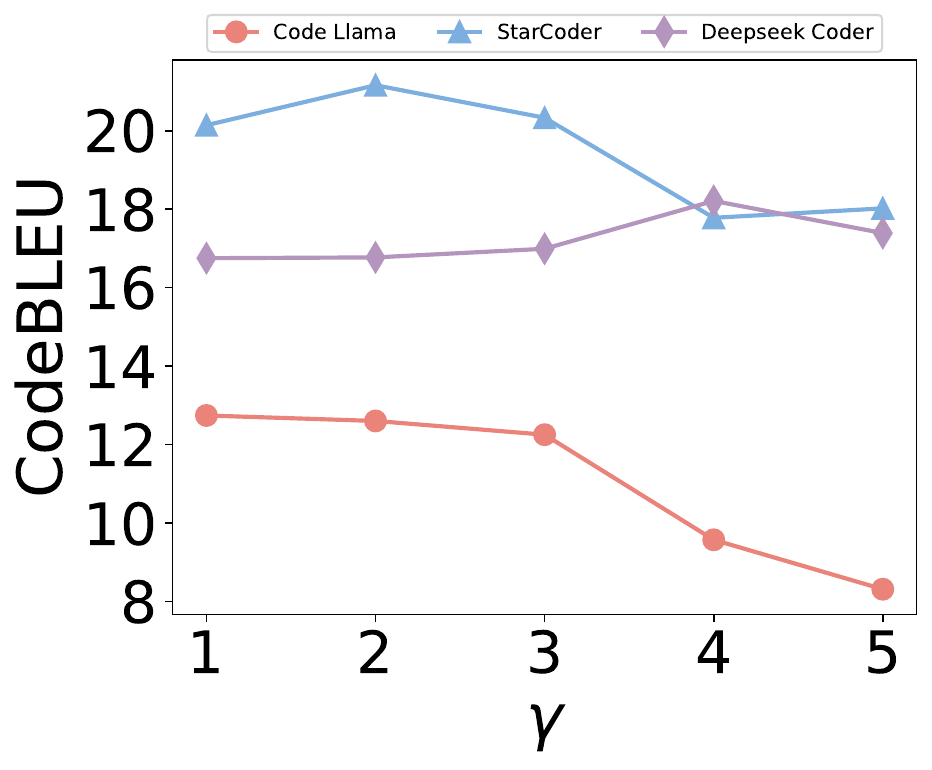}
    		\end{minipage}
		\label{fig:grid_4figs_1cap_4subcap_2}
    	}
	\subfigure[JavaScript]{
		\begin{minipage}[b]{0.23\textwidth}
			\includegraphics[width=1\textwidth]{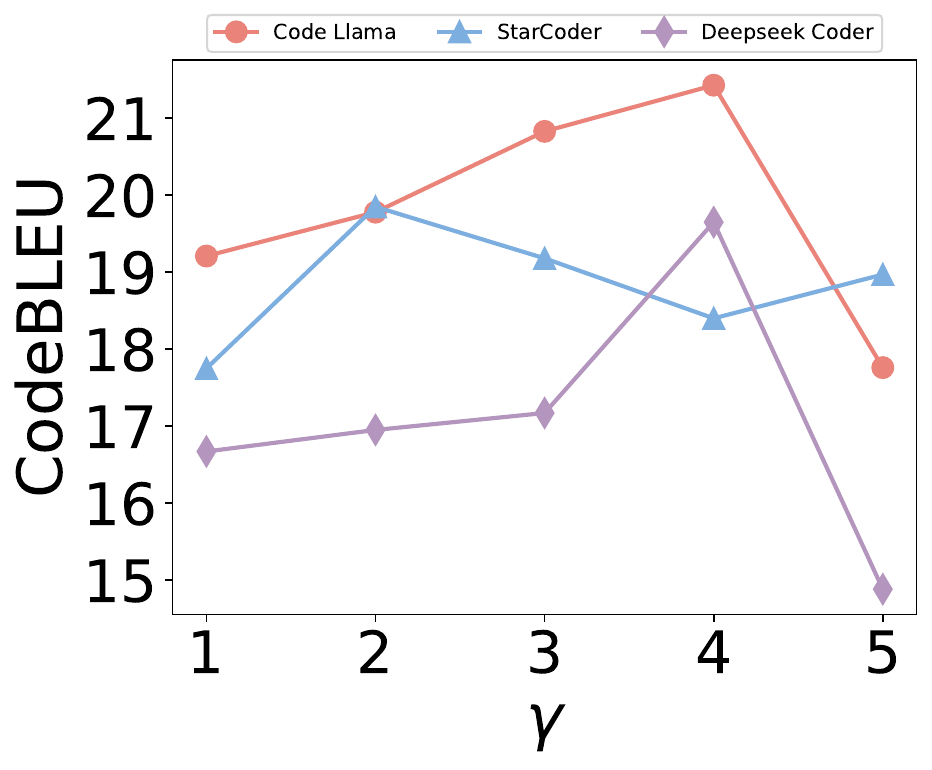} 
		\end{minipage}
		\label{fig:grid_4figs_1cap_4subcap_3}
	}
    	\subfigure[PHP]{
    		\begin{minipage}[b]{0.23\textwidth}
		 	\includegraphics[width=1\textwidth]{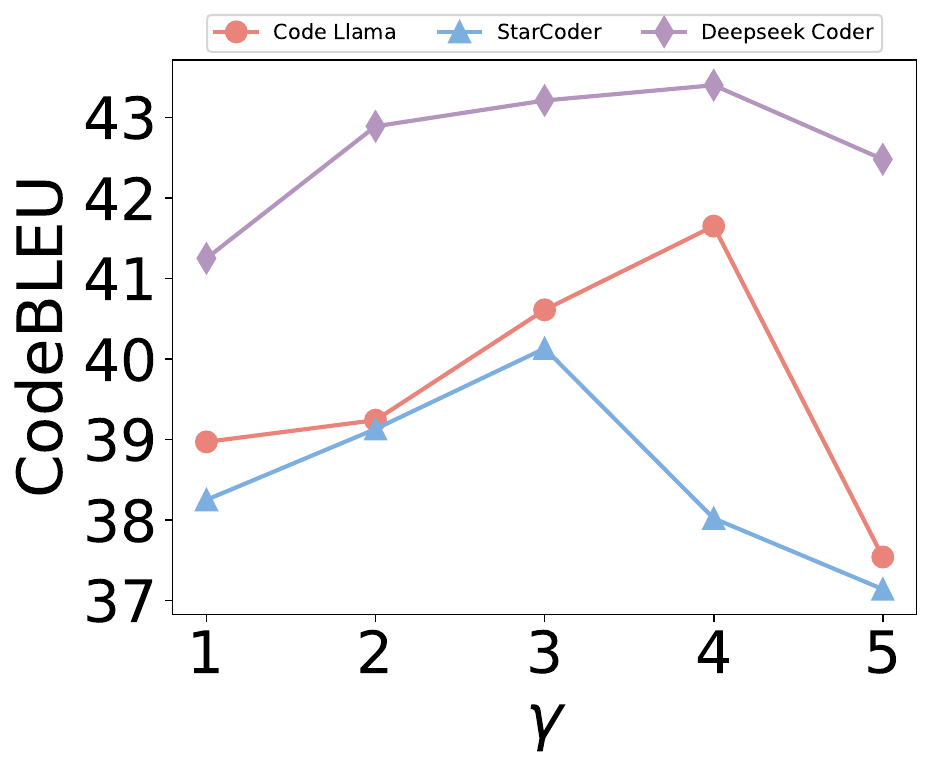}
    		\end{minipage}
		\label{fig:grid_4figs_1cap_4subcap_4}
    	}
	\caption{Impact of parameter $\gamma$ on CodeBLEU score of generated code.}
	\label{fig: app_gamma}
\end{figure*}

\begin{figure*}
	\centering
	\subfigure[Go]{
		\begin{minipage}[b]{0.23\textwidth}
			\includegraphics[width=1\textwidth]{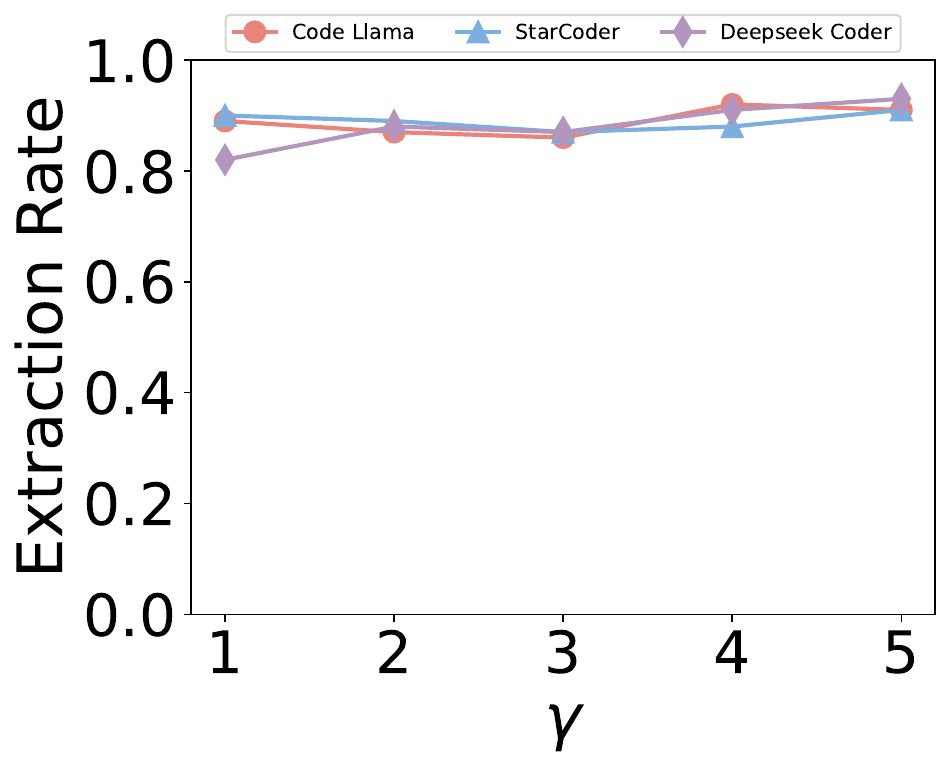} 
		\end{minipage}
		\label{fig:grid_4figs_1cap_4subcap_1}
	}
    	\subfigure[Python]{
    		\begin{minipage}[b]{0.23\textwidth}
   		 	\includegraphics[width=1\textwidth]{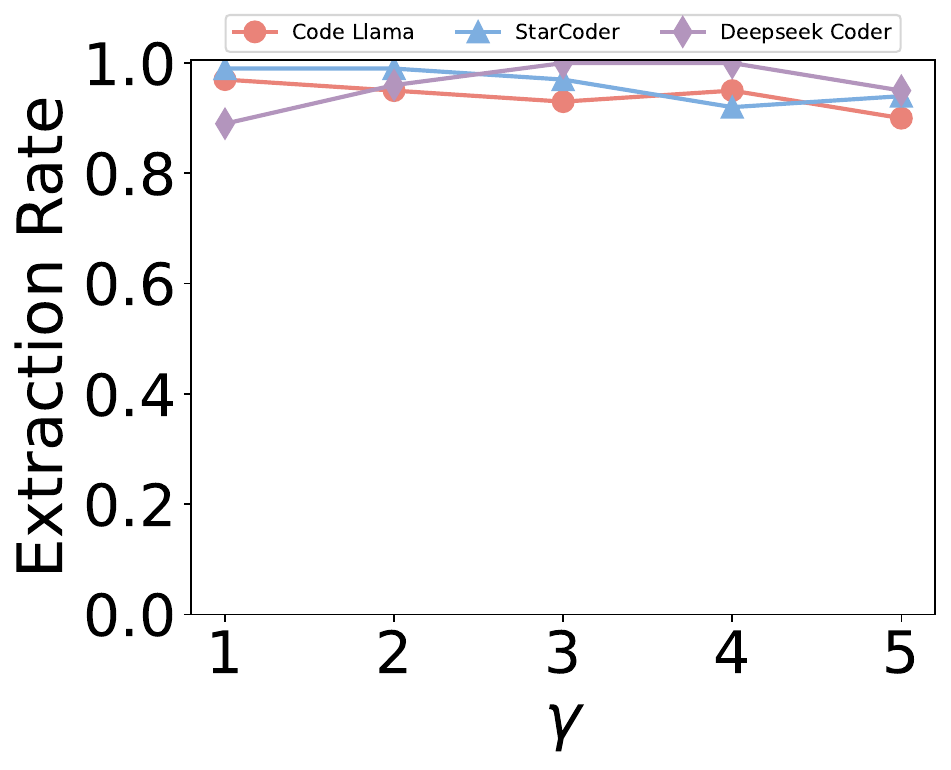}
    		\end{minipage}
		\label{fig:grid_4figs_1cap_4subcap_2}
    	}
	\subfigure[JavaScript]{
		\begin{minipage}[b]{0.23\textwidth}
			\includegraphics[width=1\textwidth]{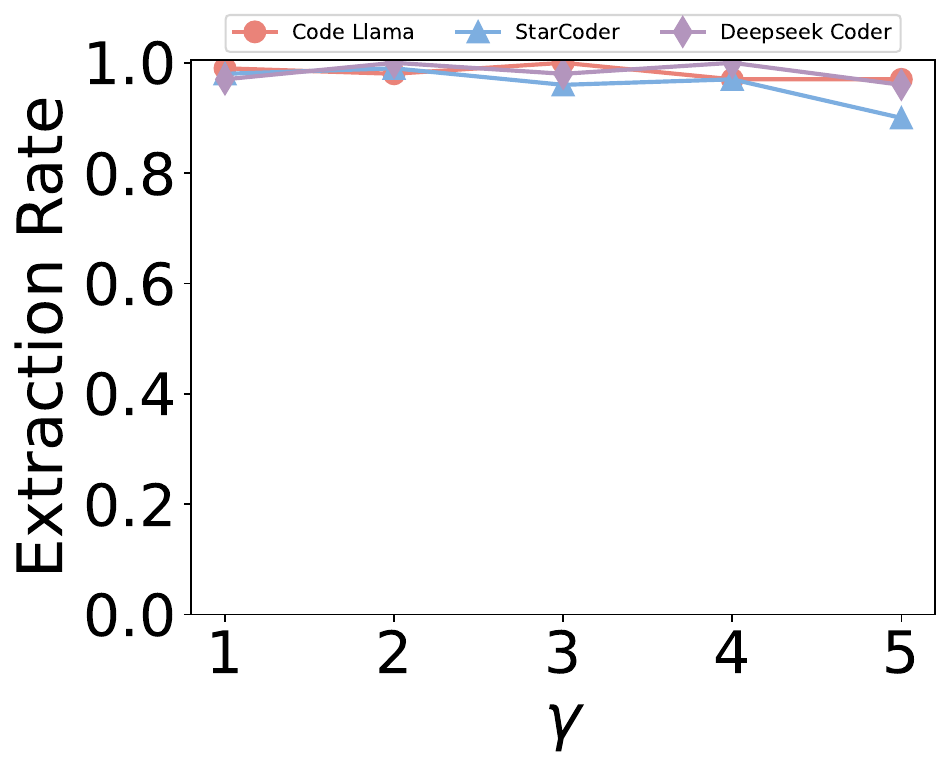} 
		\end{minipage}
		\label{fig:grid_4figs_1cap_4subcap_3}
	}
    	\subfigure[PHP]{
    		\begin{minipage}[b]{0.23\textwidth}
		 	\includegraphics[width=1\textwidth]{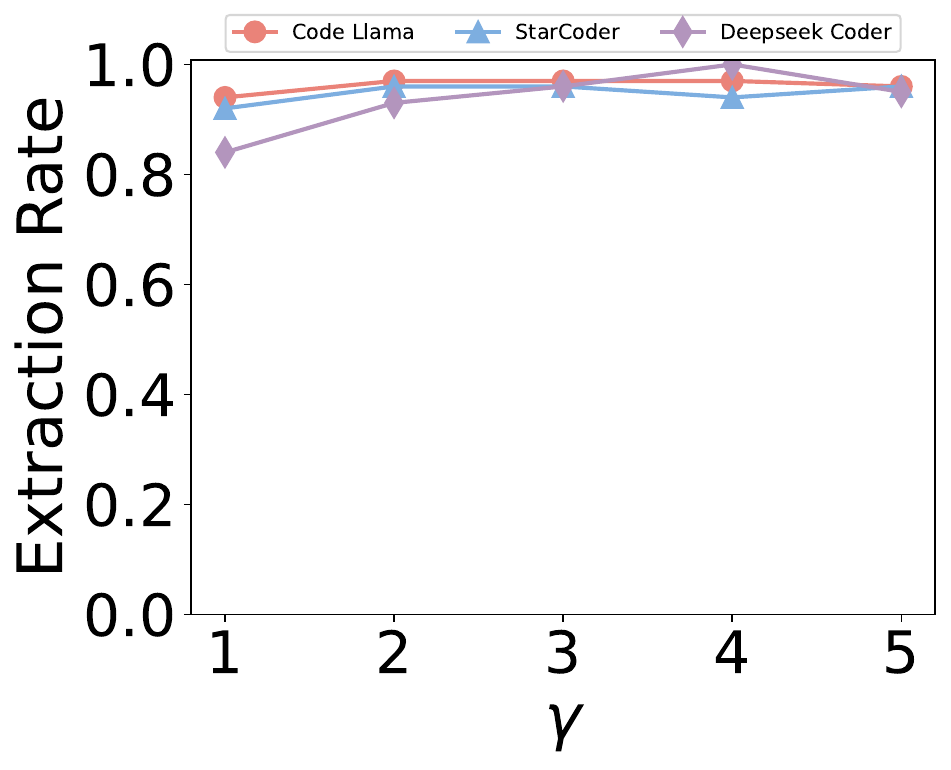}
    		\end{minipage}
		\label{fig:grid_4figs_1cap_4subcap_4}
    	}
	\caption{Impact of parameter $\gamma$ on extraction rate of generated code.}
	\label{fig: app_gaacc}
\end{figure*}

We present additional results regarding the variation of $\gamma$ with the change in CodeBLEU score and extraction rate as shown in Figure ~\ref{fig: app_gamma} and Figure~\ref{fig: app_gaacc}.

\subsection{Results of Generated Code Length}
\begin{figure*}
	\centering
	\subfigure[Go]{
		\begin{minipage}[b]{0.23\textwidth}
			\includegraphics[width=1\textwidth]{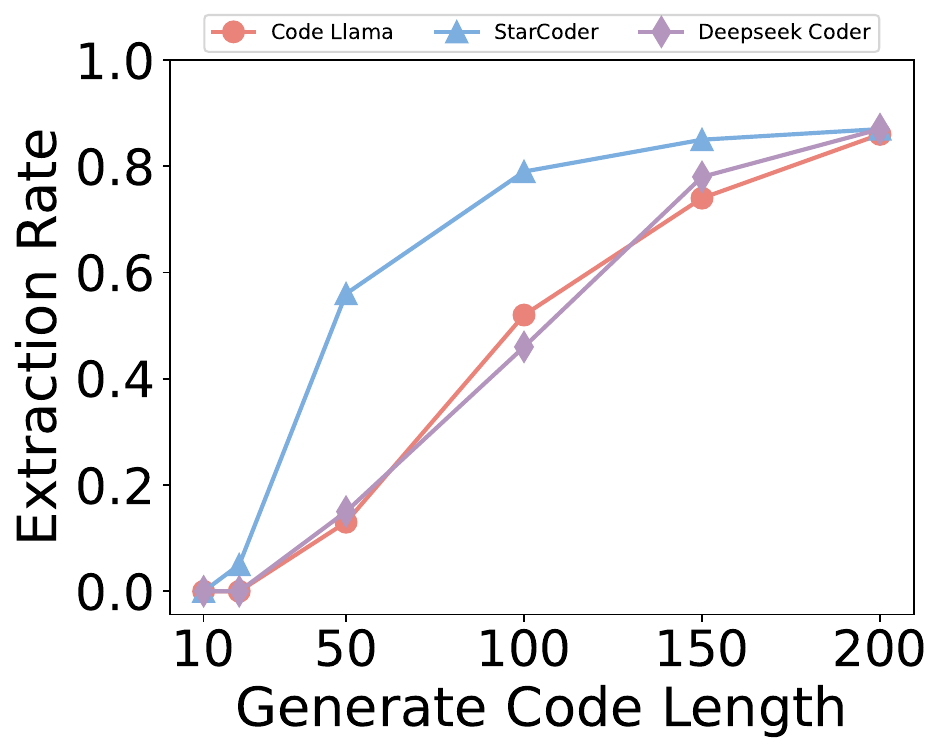} 
		\end{minipage}
		\label{fig:grid_4figs_1cap_4subcap_1}
	}
    	\subfigure[Python]{
    		\begin{minipage}[b]{0.23\textwidth}
   		 	\includegraphics[width=1\textwidth]{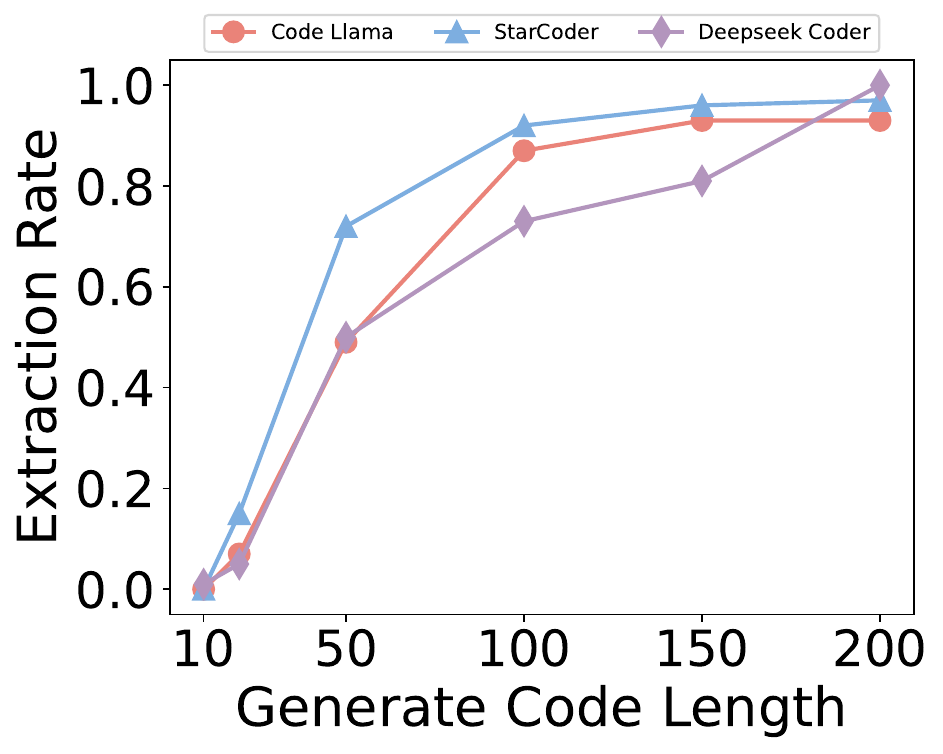}
    		\end{minipage}
		\label{fig:grid_4figs_1cap_4subcap_2}
    	}
	\subfigure[JavaScript]{
		\begin{minipage}[b]{0.23\textwidth}
			\includegraphics[width=1\textwidth]{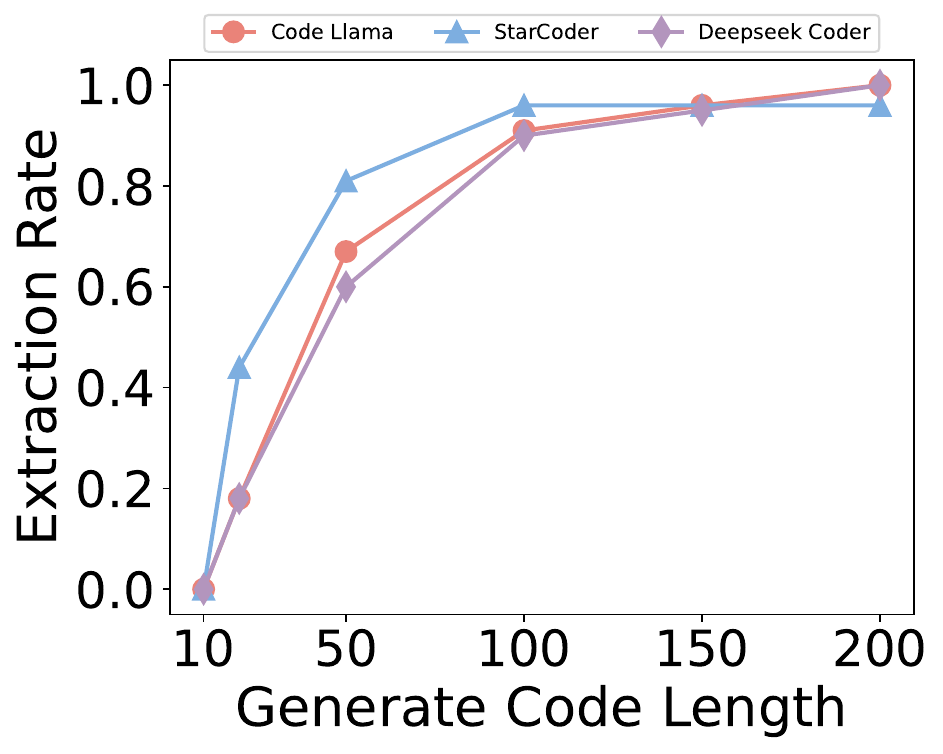} 
		\end{minipage}
		\label{fig:grid_4figs_1cap_4subcap_3}
	}
    	\subfigure[PHP]{
    		\begin{minipage}[b]{0.23\textwidth}
		 	\includegraphics[width=1\textwidth]{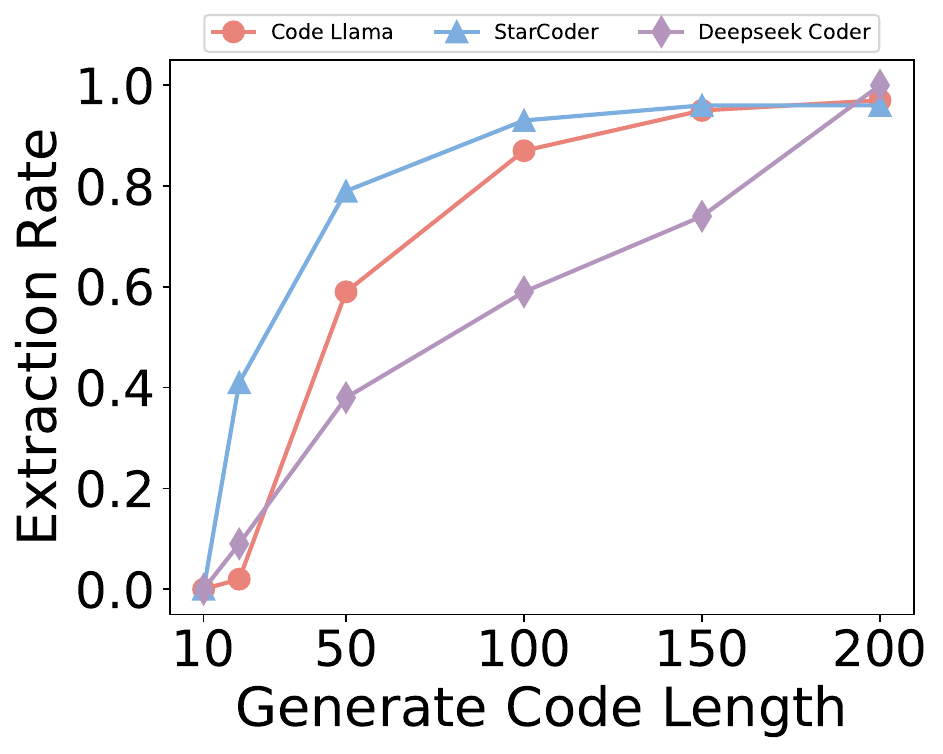}
    		\end{minipage}
		\label{fig:grid_4figs_1cap_4subcap_4}
    	}
	\caption{Impact of generated code length on the extraction rate of generated code.}
	\label{fig: app_len}
\end{figure*}

In Figure~\ref{fig: app_len}, we present additional results illustrating how the extraction rate varies with different values of generated code length.

\section{Case Study}
\begin{figure*}[!ht]
    \centering
    \subfigure[An example of Python code.
    \label{sfig: case1}]{
        
        \includegraphics[width=\textwidth]{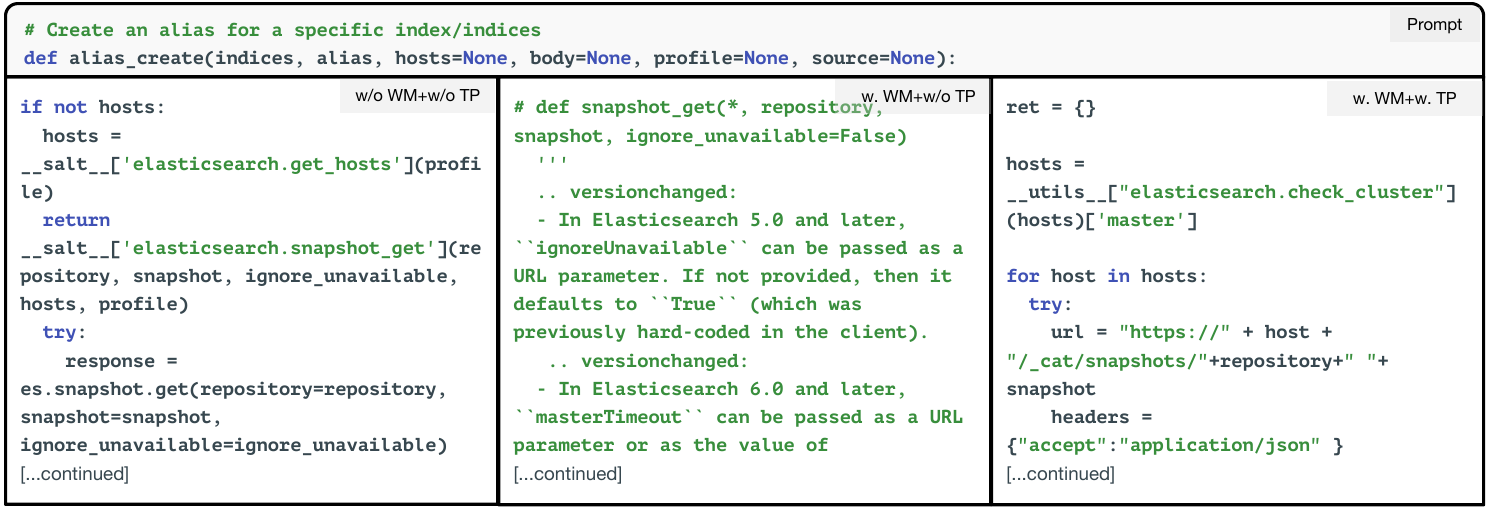}
    }
    \subfigure[An example of Go code.\label{sfig: case2}]{
        \includegraphics[width=\textwidth]{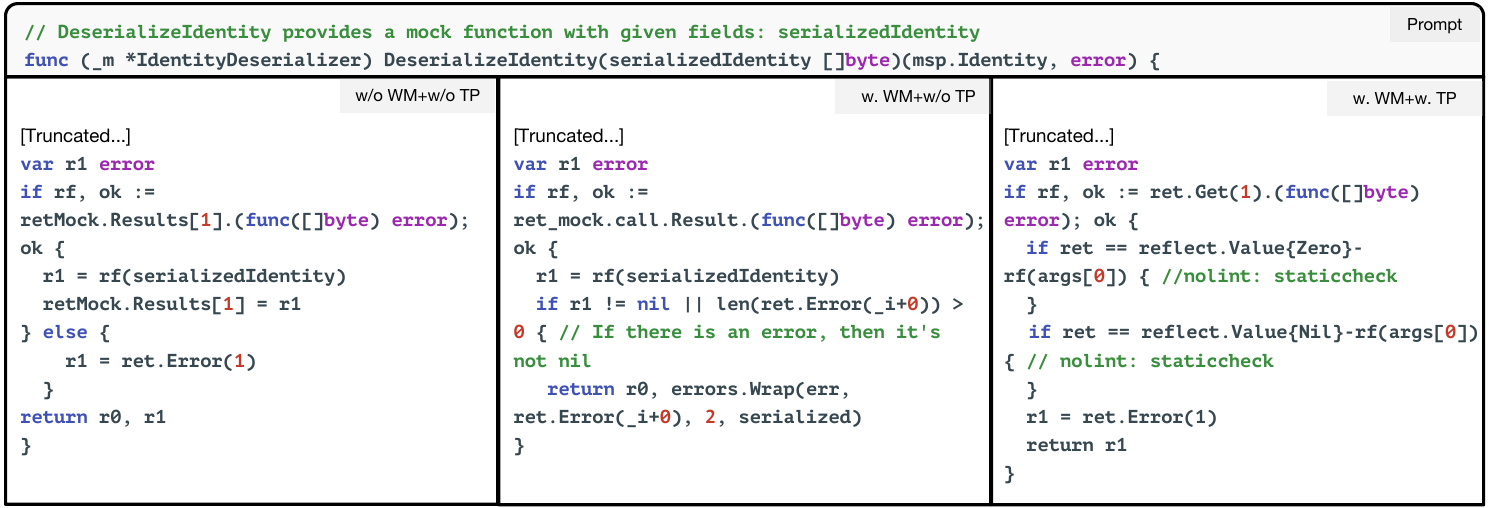}
    }
    \subfigure[An example of Java code.\label{sfig: case3}]{
	\includegraphics[width=\textwidth]{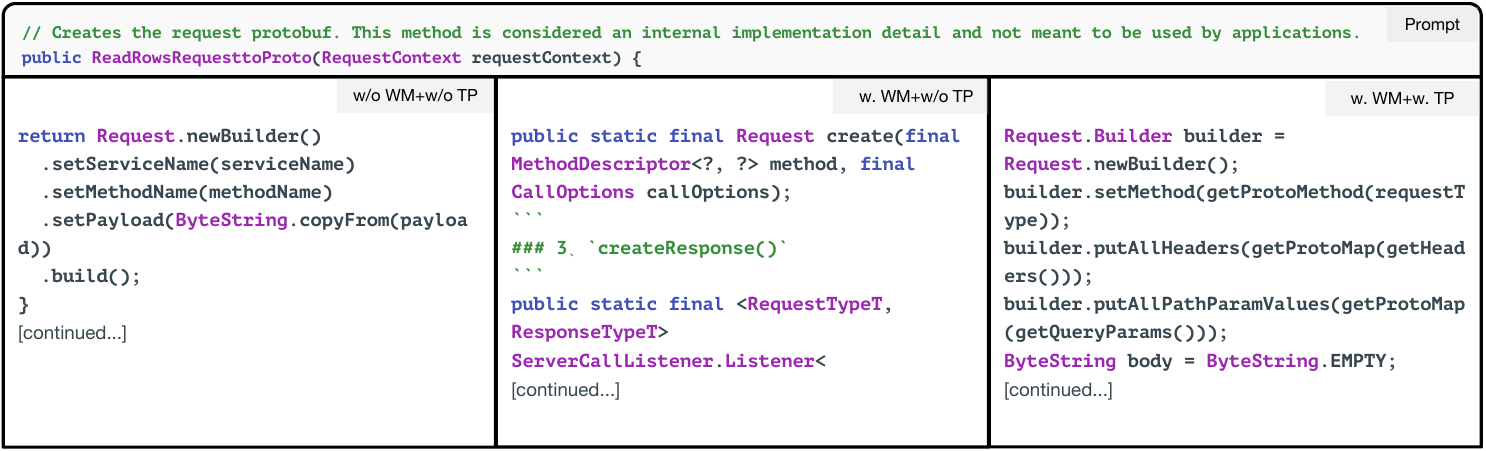}
    }
    \caption{Case study.}
    \label{fig: casestudy}
    \vspace{-1em}
\end{figure*}

In Figure \ref{fig: casestudy}, we demonstrate examples of the generation code of LLM under three different strategies(w/o WM + w/o TP, w/ WM + w/o TP, and w/ WM + w/ TP), and the watermark message is the number ``\texttt{1012}''. The prompt contains the docstring and declaration of the function.

From Figure~\ref{sfig: case1}, we can see that when watermark logit and type predictor logit are not applied during the decoding stage of LLM, it generates some normal Python code, and in this scenario, no watermark is inserted in the code because watermark logit is not applied. When only the watermark logit is added to the model logit, the LLM starts to generate large sections of comments, which is meaningless to the implementation of the function. The reason is supposed to be that the watermark logits enhance the generation probability of comment symbols like ``\texttt{\#}'' and ``\texttt{\textquotesingle\textquotesingle\textquotesingle}'', who then affect the LLM to generate comments rather than codes, which do harmness to code utility. Subsequently, when type predictor logits are also added to the model logits, the generation code of LLM resumes to normal and generates complete code to implement the function shown in the prompt.

As illustrated in Figure~\ref{sfig: case2}, the LLM generated nearly identical outputs under the three strategies. In this particular instance, no conspicuous grammatical errors were detected, and the outputs of both strategies - w/ WM + w/o TP and w/ WM + w/ TP - bear a striking resemblance to the output of w/o WM + w/o TP strategy. This case demonstrates that the watermarks we incorporated exert minimal influence on the output of LLM.

As depicted in Figure~\ref{sfig: case3}, w/ WM + w/o TP leads to meaningless generations due to the absence of grammar guidance, when w/ WM + w/ TP generates something similar to w/o WM + w/o TP, and also insert watermark ``1012'' into it. We posit that the observed outcome can be attributed to the fact that watermark logits have potentially increased the probability of erroneous type tokens being selected by LLM. Furthermore, it is discernible that once an incorrect type of token is chosen, the model's output will continually be misguided. For instance, upon the model's erroneous output of the token ``\texttt{public}'' due to the influence of watermark logits, it is anticipated that a complete function declaration will be subsequently generated by the model, thereby leading to a sustained impact on code semantics, and the generation of symbol ``\texttt{\textquotesingle\textquotesingle\textquotesingle}'' will lead to the generation of comments, which also shows the misleading effect watermark logit have on code generation task. As shown in the Figure, when type predictor logit is applied, such circumstances are unlikely to occur.

\end{document}